\documentclass[10pt,conference]{IEEEtran}
\IEEEoverridecommandlockouts
\usepackage{cite}
\usepackage{amsmath,amssymb,amsfonts}
\usepackage{graphicx}
\usepackage{textcomp}
\usepackage{xcolor}
\usepackage{subfigure}
\usepackage{multirow} 
\usepackage{booktabs}
\usepackage{url}
\usepackage[final]{changes}
\definechangesauthor[name={Dcm_replace}, color=red]{replace}
\definechangesauthor[name={Dcm_replace_img}, color=red]{replace img}
\definechangesauthor[name={Dcm_delete}, color=purple]{delete}

\newcommand\method{\textit{LogAction}\xspace}

\usepackage{algorithm}
\usepackage{algpseudocode}
\usepackage{subcaption}
\usepackage{colortbl}
\usepackage{xspace}
\usepackage{multirow}
\usepackage{enumerate}
\usepackage{colortbl}

\begin{document}

\title{LogAction: Consistent Cross-system Anomaly Detection through Logs via Active Domain Adaptation\\
}

\author{
\IEEEauthorblockN{
Chiming Duan\thanks{\IEEEauthorrefmark{10} Work was done when Chiming was an intern at Bytedance.}\IEEEauthorrefmark{10}\IEEEauthorrefmark{2}\IEEEauthorrefmark{3},
Minghua He\thanks{\IEEEauthorrefmark{3} Equally Contribution.}\IEEEauthorrefmark{2}\IEEEauthorrefmark{3},
Pei Xiao\IEEEauthorrefmark{2}\IEEEauthorrefmark{3},
Tong Jia\IEEEauthorrefmark{4}\IEEEauthorrefmark{6}\IEEEauthorrefmark{1},
Xin Zhang\IEEEauthorrefmark{8},
Zhewei Zhong\IEEEauthorrefmark{8},
Xiang Luo\IEEEauthorrefmark{8},\\
Yan Niu\IEEEauthorrefmark{8},
Lingzhe Zhang\IEEEauthorrefmark{2},
Siyu Yu\IEEEauthorrefmark{2},
Yifan Wu\IEEEauthorrefmark{2},
Weijie Hong\IEEEauthorrefmark{2},
Ying Li\IEEEauthorrefmark{2}\IEEEauthorrefmark{5}\IEEEauthorrefmark{1}\thanks{\IEEEauthorrefmark{1} Corresponding author.}
and Gang Huang\IEEEauthorrefmark{6}}
\IEEEauthorblockA{\IEEEauthorrefmark{2}School of Software and Microelectronics, Peking University,
Beijing, China \\
\{duanchiming, hemh2120, xiaopei, zhang.lingzhe, hongwj\}@stu.pku.edu.cn, gaiusyu6@gmail.com, \{yifanwu, li.ying\}@pku.edu.cn
}

\IEEEauthorblockA{\IEEEauthorrefmark{4}Institute for Artificial Intelligence, Peking University,
Beijing, China \\
jia.tong@pku.edu.cn}

\IEEEauthorblockA{\IEEEauthorrefmark{5}National Engineering Research Center for Software Engineering, Peking University,
Beijing, China \\
li.ying@pku.edu.cn}

\IEEEauthorblockA{\IEEEauthorrefmark{6}National Key Laboratory of Data Space Technology and System, Beijing, China \\
\{jia.tong, hg\}@pku.edu.cn}

\IEEEauthorblockA{\IEEEauthorrefmark{8}Bytedance, Beijing, China \\
\{zhangxin.11, zhongzhewei, luoxiang.10, niuyan.13\}@bytedance.com}

}

\maketitle

\begin{abstract}
Log-based anomaly detection is a essential task for
ensuring the reliability and performance of software systems. However, the performance of existing anomaly detection methods heavily relies on labeling, while labeling a large volume of logs is highly challenging. To address this issue, many approaches based on transfer learning and active learning have been proposed. Nevertheless, their effectiveness is hindered by issues such as the gap between source and target system data distributions and cold-start problems. In this paper, we propose \method, a novel log-based anomaly detection model based on active domain adaptation. \method integrates transfer learning and active learning techniques. On one hand, it uses labeled data from a mature system to train a base model, mitigating the cold-start issue in active learning. On the other hand, \method utilize free energy-based sampling and uncertainty-based sampling to select logs located at the distribution boundaries for manual labeling, thus addresses the data distribution gap in transfer learning with minimal human labeling efforts. Experimental results on six different combinations of datasets demonstrate that \method achieves an average 93.01\% F1 score with only 2\% of manual labels, outperforming some state-of-the-art methods by 26.28\%. Website: \url{https://logaction.github.io}
\end{abstract}


\section{Introduction}
\label{sec:intro}
Software systems are becoming increasingly large and complex and are subject to more failures. 
As system logs record system states and significant events of running processes, they are an excellent source of information for anomaly detection. 
Log-based anomaly detection is promising for system reliability and has been widely studied.
% Microservices architecture have gained immense popularity owing to their capacity to foster agility, scalability, fault tolerance and diverse technology stacks. However, the inherent complexity and distributed nature of microservices architecture can pose challenges for effective anomaly detection due to increased monitoring intricacies and diverse failure points. Logs, by capturing system states and operational events, serve as critical data sources for detecting anomalies within these architectures. Leveraging log-based anomaly detection holds significant promise for enhancing system reliability and has garnered extensive research attention.

Existing log-based anomaly detection models can be broadly divided into two categories: unsupervised models and supervised models.
Unsupervised models ~\cite{deeplog,loganomaly,icsme2020,icse2020,he2023unsupervised,llmelog,weakly-supervised-logad,clslog} utilize sequential neural networks such as LSTM, GRU, etc. to learn the occurrence possibility of log events in normal event sequences to predict subsequent log events and identify the unmatched log event as an anomaly.
Their effectiveness is very limited due to the lack of supervision of anomalous logs ~\cite{plelog}.
Supervised models ~\cite{logrobust,bugidentifier,recurrentfaults,xie2023logrep} build classification models to identify anomalous logs and are more effective compared to unsupervised models. 
However, their effectiveness heavily relies on large amounts of labeled logs.

In real-world software systems, identifying anomalous logs poses a significant challenge due to the vast volume of system logs in which such anomalies are deeply buried.
Consequently, obtaining accurate data labels, especially for anomalous logs, is a rare and difficult task ~\cite{hilog}. To solve this problem, two different ideas are proposed. First, transferring abundant historical labeled logs from other mature systems (source systems) to new systems (target systems) with a few labeled logs for model training, namely transfer-learning-based methods~\cite{logtransfer,logtad}. 
%These methods xx.
% These methods take the assumption that despite differences in log formats and content across various systems, they have similar semantics. They utilize word embedding models to extract semantic information from raw logs and organize it into log sequences. 
These methods first use the log sequences from the source systems to train an anomaly-detection model and then fine-tune it using log sequences from the target systems. 
These works have two main issues. First, their applicability is confined to scenarios where the data distribution gap of log sequences in source and target systems is small. For instance, LogTransfer~\cite{logtransfer} can only handle scenarios that the source system and target system belong to the same type of switch logs or software family (e.g., Hadoop application and Hadoop file system) which severely limit its availability. 
% Because of the variations in the structures and fault-tolerance mechanisms, different systems exhibit differing frequencies and types of anomalies. In other words, the data distribution gaps in log sequences from different system remain significant, despite semantic similarities. 
Second, their utilization of labels from the new system is inadequate. For instance, MetaLog~\cite{metalog} utilizes a meta-learning paradigm
to enhance the model’s generalization ability, constructing
anomaly detection models for new systems.
However, due to the lack of effective label utilization, substantial human labeling efforts is required to learn the complete data distribution in the new system. 
% Conversely, LogTAD~\cite{logtad} employs an unsupervised approach by transferring the hyper-sphere centers to achieve model transfer. It has been demonstrated that without human labels, the model's efficacy is severely limited ~\cite{plelog,hilog}. 
% In fact, acquiring labels from a new system is feasible, albeit to minimize human labeling efforts, it is preferable to select the most useful subset~\cite{EADA};

Second, automatically and actively selecting the most important logs for humans to label, namely active-learning-based methods~\cite{aclog,afalog,eagerlog,logcae}. 
%These works leverage active learning xx. 
These works leverage active learning, enabling models to actively select samples for human annotations, in order to achieve highly efficient anomaly detection models with minimal human labeling efforts. However, existing active-learning-enabled works suffer from a cold start problem. 
% First, the precious human labels are not fully used by unsupervised backbone models. These works leverage unsupervised models as backbones and use human labels simply to reduce noises in the training datasets. 
% %For instance, ACLog xx.
% For instance, AcLog~\cite{aclog} selects “fuzzy” logs for human annotation and filters noisy logs from the training set by computing the Levenshtein distance between the logs in the training set and the labeled anomalous logs, then it retrains the LSTM-based unsupervised model with the noise-filtered training set. 
% However, if a labeled anomalous log doesn't offer any useful information to filter out noise, the labeling effort for that log becomes nonsensical. 
% As a result, the precious human labels are not fully used. 
% As a result, the precious labeled anomalous logs are simply used for filtering tnot exactly learned by the model
% In fact, the unsupervised backbone models can not learn the features of labeled anomalous logs and still suffer from the lack of supervision leading to unsatisfied performance. 
% Second, existing works suffer from a cold start problem. 
This implies that the model's ability is heavily dependent on the accumulation of online human labels, particularly those associated with anomalous logs. 
For instance, ACLog ~\cite{aclog} necessitates 205 online labels of logs from the BGL dataset ~\cite{oliner2007supercomputers}. 
Accumulating a sufficient number of labels requires a significant amount of time, during which the model's performance remains limited. 

In our view, transfer-learning-based methods aim to increase the amount of labeled data by leveraging external labeled data from other systems, while active-learning-based methods aim to increase the quality of labeled data by carefully selecting data instances to be labeled. They consider solving the label lacking problem from different aspects and can benefit each other. From the perspective of transfer-learning-based methods, active learning can help use data labels of key data instances to bridge the gap between different systems. From the perspective of active-learning-based methods, transfer learning can help solve the \replaced[id=replace]{cold}{code} start problem. 
% In our view, transfer-learning-based methods effectively leverage historical labeled logs but struggle with identifying new, useful logs for labeling that could reduce the distribution gaps. Conversely, active-learning-based methods excel at acquiring new labeled logs that is highly valuable but often suffer from the cold-start problem. Both of them can reduce the human labeling efforts from different perspectives but suffer from huge data gap and cold start problem separately. 

\begin{figure}[htbp]
\centerline{\includegraphics[width=\linewidth]{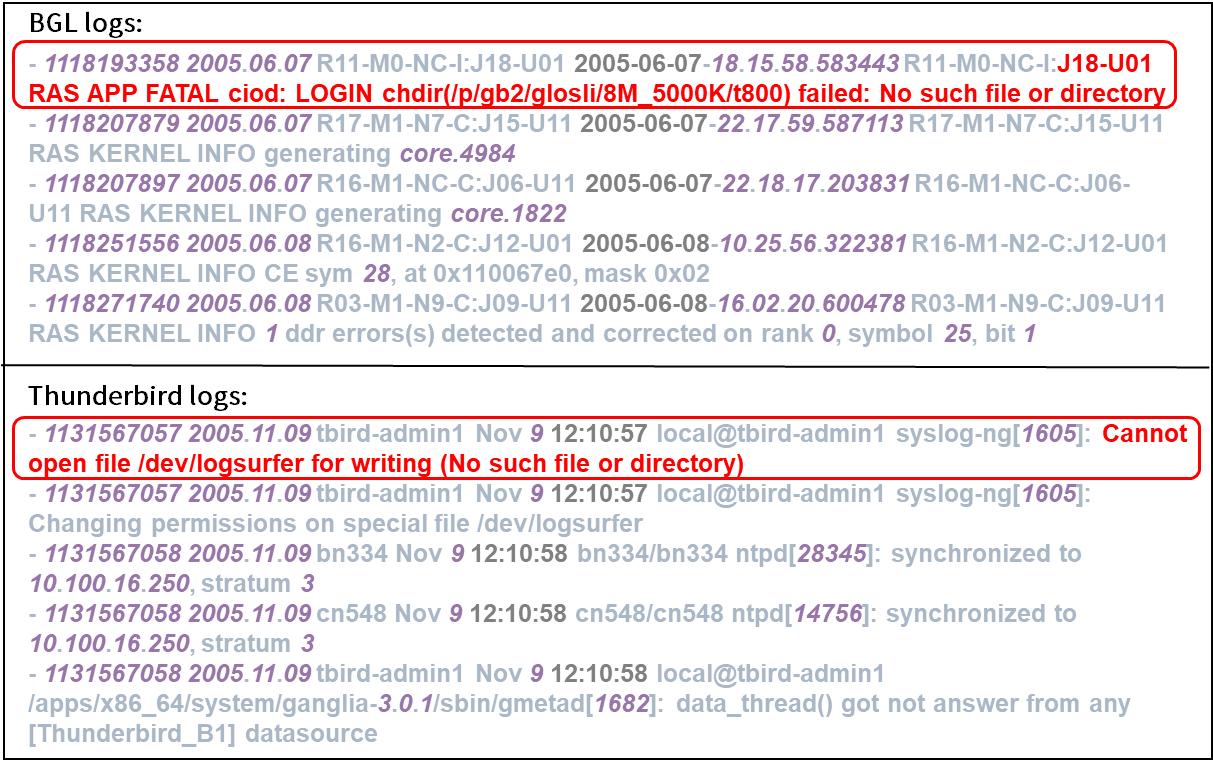}}
\caption{Two log sequences from different systems (BGL and ThunderBird). Although they express the same error - file or directory does not exist, their formats show distinct differences.}
\label{fig:logs}
\vspace{-0.4cm}
\end{figure}

As a result, we pose an idea that transfer learning and active learning should be combined together to solve the label lacking problem. We define this scenario as consistent cross-system anomaly detection (CCAD), that is, leveraging the features extracted from abundant historical labeled logs of mature systems (source systems) to build anomaly detection models for new systems (target systems) and consistently optimize the models with online human labels on the target systems.  
In this paper, we focus on the CCAD scenario and aim to build a high performance anomaly detection model without or with very few anomalous labels. 

However, achieving this is not easy. First, bridging the huge data distribution gaps between source system and target systems is challenging. 
%xxx.
Logs generated by different systems exhibit varying formats and content. As illustrated in Figure \ref{fig:logs}, logs originating from two distinct service systems both convey the same error, 'file not exist,' yet they significantly differ in their log formats and content. Moreover, the types and frequencies of anomalies vary across different systems, further amplifying the data distribution gaps. 
Second, accurately selecting the least but most useful logs for human labeling from huge amount of system logs is challenging. 
%xxx.
Throughout system operations, thousands of logs can be generated per second. The human effort to label every log is remarkably high. Furthermore, these logs frequently exhibit substantial redundancy, with the information contained within this redundancy often tracing back to previously labeled logs. So, the selection of the most useful logs for human labeling is crucial. These logs should encompass distributional information that hasn't been captured in any historically labeled logs, aiming to minimize the human labeling efforts.
Facing these challenges, in this paper, inspired by ~\cite{EADA}, we propose 
%xxx.
\method, a consistent cross-system anomaly detection model via active domain adaptation. Active domain adaptation, a fusion of active learning and transfer learning techniques, has demonstrated its effectiveness in various studies, particularly in reducing the human labeling efforts~\cite{rai2010domain,EADA,han2022loss,fu2021transferable}.

To handle the first challenge, we use global embedding and contrastive learning. We employ a pretrained BART model~\cite{BART} to extract meaning from diverse log formats, converting them into word vectors in a unified space. These word vectors are then turned into log sequences using time windows. Subsequently, we utilize contrastive learning to encode log sequences from diverse systems into log vectors with similar distributions, effectively minimizing the data distribution gaps across different systems. This allows knowledge from the source system to be applied to the target system, enabling us to create a base anomaly detection model for the target using the source's log vectors.

For the second challenge, we turn to active learning. This helps us select the most informative yet fewest log vectors in the target system for human labeling to improve the base model. We employ two sampling methods, free energy-based sampling and uncertainty-based sampling, to collect log vectors that are distinct from both historically labeled source log vectors and those previously labeled during selection. In our view, these log vectors holds the most informative content. On the one hand, these log vectors contain information the model hasn't encountered before, making them more valuable for the model's learning. On the other hand, these log vectors, distinct from those of the source system, encapsulate information regarding the gap in data distribution between the source and target systems. Learning from this subset of log vectors facilitates further mitigation of the distributional gaps between the source and target systems. 
We validated the efficacy of \method across three distinct public datasets, each representing a completely different system. The experiments demonstrate that our approach surpasses 93\% F1-score performance with the utilization of only 2\% labeled data in target system. Besides, our approach outperforms state-of-art models that can use for CCAD task by 26.28\% on average. 
The contribution of this paper are as follows:
\begin{enumerate}
    \item We propose \method, a novel generalizable anomaly detection approach via active domain adaptation.
    \item We utilize global embedding and contrastive learning techniques to mitigate the data distribution gap, while employing active learning to reduce human labeling efforts, effectively addressing the two primary challenges in CCAD task.
    \item Evaluation results on three open datasets show the significant effectiveness of our approach.
\end{enumerate}

\section{Related Work}
\label{sec:relate_work}

\subsection{Log-based anomaly detection}
Timely anomaly detection plays a vital role in ensuring system reliability\cite{AgentFM,enhancing-web,zhang2025thinkflselfrefiningfailurelocalization,FAMOS,soil,zhang2025adaptivercl}. Analyzing logs for problem detection and identification has been an active research area ~\cite{deeplog,loganomaly,logrobust,icsme2020,icse2020,logflash,cloudseer,kddcfg,logsed,icwslogsed,logdc,ava,beehive,logan}. These work first parse logs into log events, and then build anomaly detection models. Some state-of-art approaches ~\cite{logflash,cloudseer,kddcfg,logsed,icwslogsed,logdc,ava,beehive,logan} extract event sequence at first, and then generate a graph-based model to compare with log sequences in production environment to detect conflicts. Other approaches often build deep learning-based models ~\cite{deeplog,loganomaly,logrobust,icsme2020,icse2020,he2023unsupervised,xie2023logrep} to capture the sequence features of log events. Deeplog is a typical work ~\cite{deeplog} that utilizes LSTM network ~\cite{lstm} to model the sequence of log events and the sequence of variables in log text. LogAnomaly ~\cite{loganomaly} utilizes a word2vec model to transform events into vectors with semantic features to improve the anomaly detection result. LogRobust ~\cite{logrobust} utilizes TF-IDF and word vectorization to transform logs into semantic vectors. In this way, updated new logs can be transformed into semantic vectors and participate in model training and deduction. LogTransfer~\cite{logtransfer} and LogTAD~\cite{logtad} use transfer learning to achieve the cross-system anomaly detection.
MetaLog~\cite{metalog} use meta-learning paradigm
to enhance the model’s generalization ability, constructing anomaly detection models for new systems. However, their effectiveness is constrained by the data distribution gap between domains and the insufficient utilization of labels. As a result, they are not suitable for addressing the CCAD task.

\subsection{Active domain adaptation}
Active domain adaptation (ADA) combines the principles of active learning and transfer learning, enabling a model to actively select the valueable samples during transfer to the target domain, thereby reducing the need for human labeling efforts. Rai et al.~\cite{rai2010domain} were the first to demonstrate the synergistic relationship between active learning and transfer learning, prompting various related works. Recently, AADA~\cite{su2020active} has employed adversarial training and ADA to address domain shift issues. Bo et al.~\cite{fu2021transferable} utilized Transferable Query Selection (TQS) for sample selection, aiming to overcome the limitations of single-domain active learning in ADA. Han et al.~\cite{han2022loss} improved the performance of active domain adaptation by employing loss-based querying for sample selection. EADA~\cite{EADA} employed an energy-based model for sampling, considering both sample entropy and uncertainty. These approaches have exhibited promising performance in tasks such as image classification and successfully inspire us to integrate active learning and transfer learning to address the CCAD tasks.
\section{Approach}
\label{sec:approach}
In this paper, we introduce 
\method, an innovative approach based on active domain adaptation, specifically designed to address the two key challenges in CCAD tasks as defined in Section \ref{sec:intro}. Existing works often inadequately address these challenges, resulting in their subpar performance. In contrast, after log parsing, \method utilizes global embedding and contrastive learning to map the log sequences of the source and target systems into similar distributions, aiming to mitigate the data distribution gaps. Then, \method leverages a substantial amount of labeled data from the source system to train the anomaly detection model and fine-tunes it on a high-information subset selected through active learning. In this way, we can obtain a generalized model and make it applicable to anomaly detection tasks in the target system with minimal human labeling efforts. In this section, we will introduce our model. In Section \ref{sec:Overview}, we provided an overview of our model. In Sections \ref{sec:Encoding} and \ref{sec:ActiveDomainAdaptation}, we will separately discuss the two primary components of our model: Encoding and Active Domain Adaptation.
% We can obtain a generalized model and make it applicable to anomaly detection tasks in the target system with minimal human labeling costs.
\subsection{Overview}
\label{sec:Overview}
\begin{figure*}[htbp]
\small
\centerline{\includegraphics[width=\textwidth]{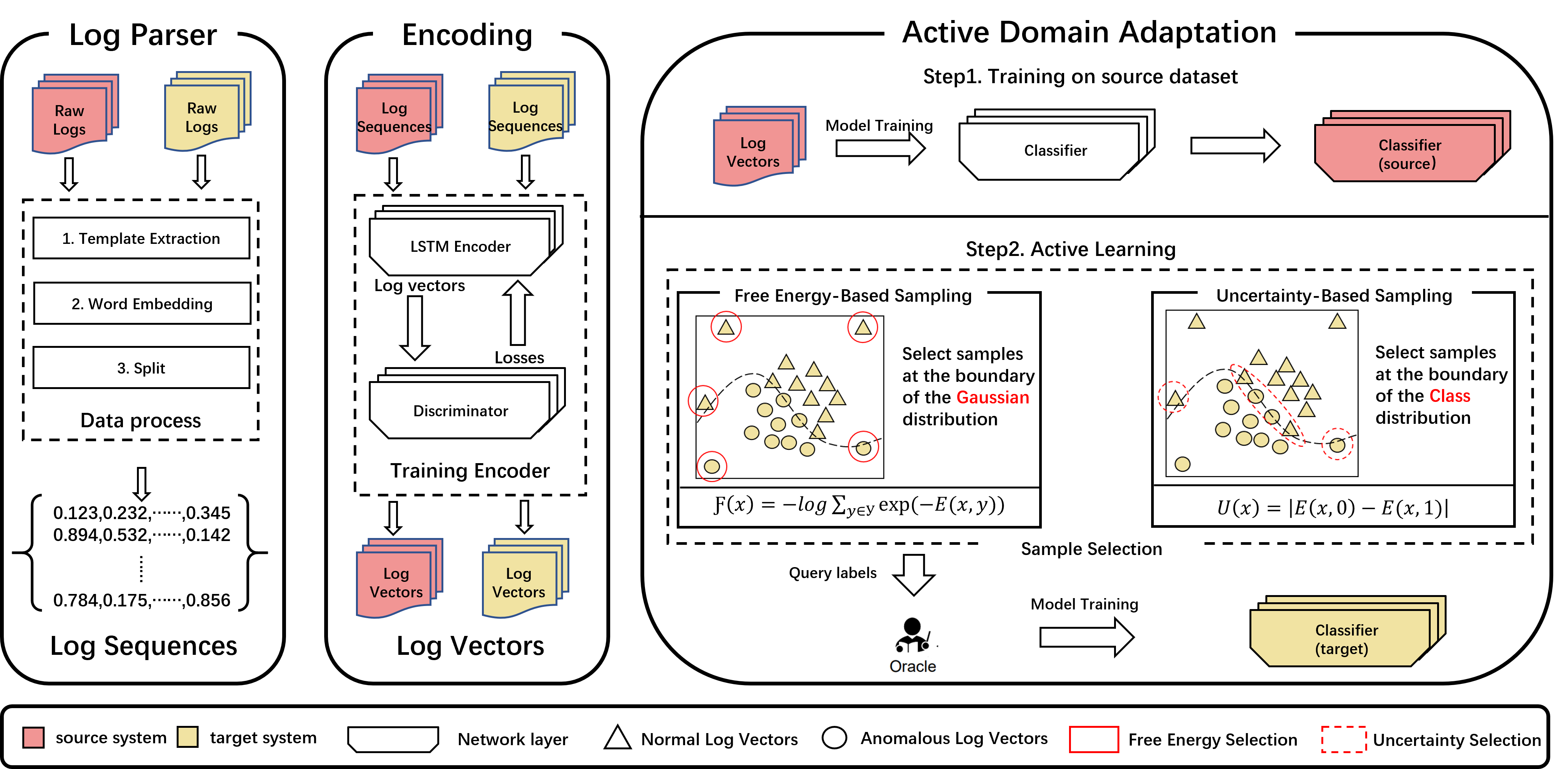}}
\caption{The overview of \method. 
\method includes three main phases: Log Parser, Encoding and Active Domain Adaptation. 
Firstly, the raw system logs are labeled and parsed into log event sequences. 
Secondly, \method encodes the log sequences from both the source system and the target system, mapping them to similar distribution. 
Finally, \method is initially trained using labeled log vectors from the source system. Subsequently, it is fine-tuned with a very limited amount of target system logs via active learning to adapt to cross-system anomaly detection.
}
\label{fig:overview}
\end{figure*}

% System's logs consist of lines of text messages, where each line represents a log event containing system status information at the current timestamp. Our approach for anomaly detection relies on analyzing multiple log events within a time window, as they collectively encapsulate system state information for that specific time interval. We refer to this collection of log events within the time window as a log sequence.
Figure \ref{fig:overview} illustrates the overview of the \method. \method includes three main phases: Log Parser, Encoding and Active Domain Adaptation. 
In the Log Parser phase, we employ the advanced log parsing method Drain~\cite{drain} to obtain templates of log events. Subsequently, we utilize the pre-trained BART model~\cite{BART} to extract the semantic information from these log event templates, transforming them into word vectors. Pre-trained models have demonstrated superior semantic extraction capabilities compared to word embedding models like GloVe~\cite{Glove} and Word2Vec~\cite{word2vec}. Afterward, we employ a sliding time window to split the log event templates represented by word vectors, forming the log sequences. As illustrated in Figure \ref{fig:overview}, each log sequence is a two-dimensional matrix, where each row represents a word vector of a log event.
In the Encoding phase, we encode the log sequences from both the source system and the target system, mapping them to similar distribution. The resulting encoded log sequences are referred to as log vectors. We utilize contrastive learning to train the encoder, treating normal log sequences from both the source system and the target system as one class and anomalous log sequences as another. 
\added{During the \method process, the encoder is trained jointly with the downstream anomaly detection model, and the labels come from active learning processes.}
A detailed description of our encoder implementation will be provided in Section \ref{sec:Encoding}.
In the Active Domain Adaptation phase, we initially use labeled log vectors from the source system to train an anomaly detection model called the "Classifier(source)". Because it is trained on log vectors from a source system with similar distributions to target system, the Classifier(source) possesses a certain degree of generalization. However, to further tailor it for the target system, we require some labeled data from the target system to fine-tune it. To minimize the need for human labeling during fine-tuning, we utilize active domain adaptation techniques to accomplish this phase. We primarily employ two modes of sampling using active learning: free energy-based sampling and uncertainty-based sampling. In the free energy-based sampling phase, we choose log vectors with the highest free energy deviation from the source system. In the uncertainty-based sampling phase, we select log vectors situated at the boundary between normal and anomalous log vector classifications. 
Subsequently, we manually label the log vectors chosen through active learning and fine-tune Classifier(source) into Classifier(target) using them to complete the domain adaptation process. Detailed insights into our Active Domain Adaptation approach will be provided in Section \ref{sec:ActiveDomainAdaptation}.

\begin{figure}[htbp]
\centerline{\includegraphics[width=\linewidth]{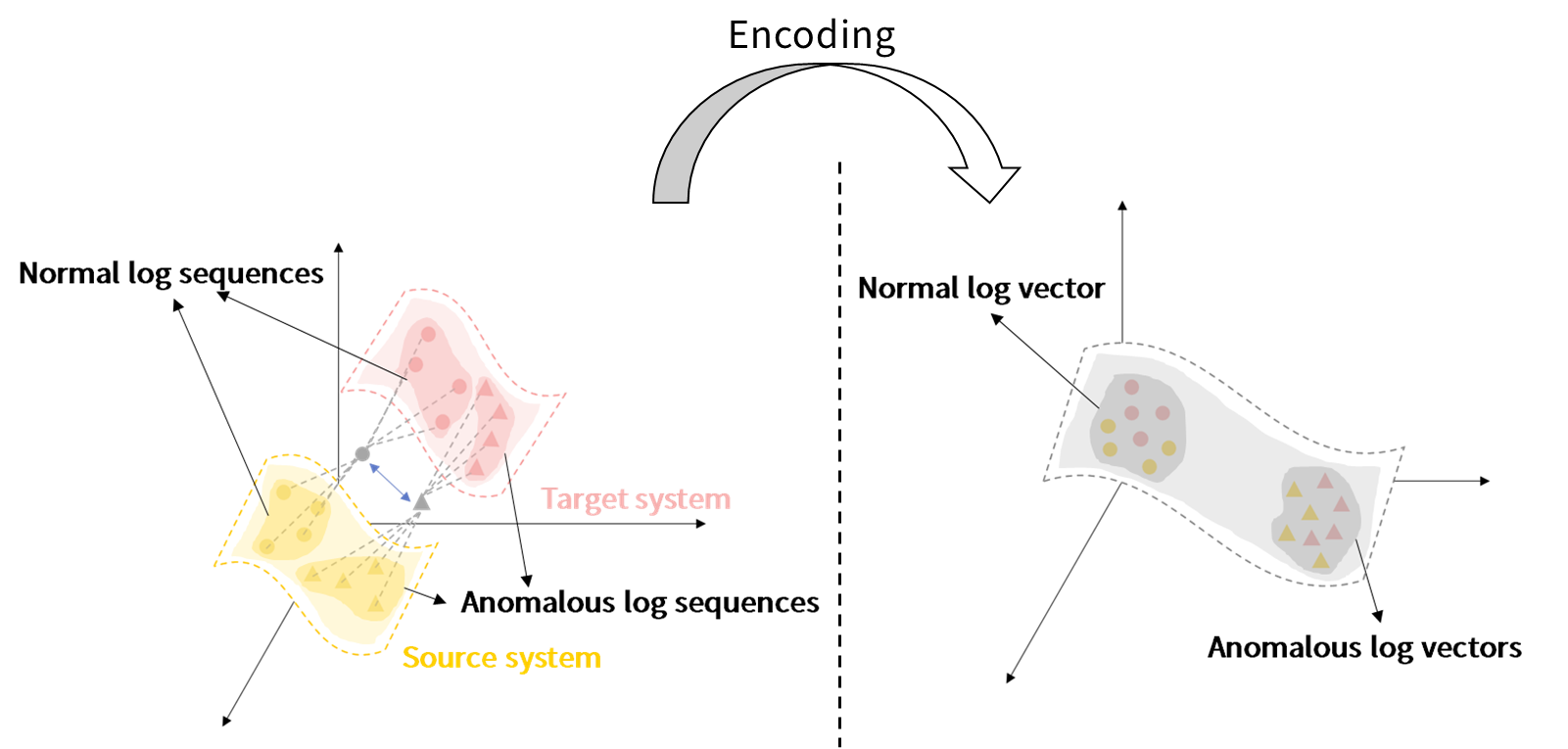}}
\caption{The Encoding Phase. In the initial stage, log sequences from the source system and the target system originate from distinct distributions (located in different hyperplanes). Employing contrastive learning, the objective is to map the distributions of normal log sequences from the source system and normal as well as anomalous log sequences from the target system to similar distributions (situated within the same hyperplane).}
\label{fig:contrastive learning}
\vspace{-0.4cm}
\end{figure}

\subsection{Encoding}
\label{sec:Encoding}
After log parsing, diverse formats of raw logs are parsed into log sequences within the same feature space. However, the fault-tolerance mechanisms and structural differences contribute significantly to variations in error categories and frequencies across different systems. As a result, significant data distribution gaps still exist in the log sequences between the source system and the target system. So, we utilize contrastive learning to map the log sequences from the source system and the target system into log vectors, aligning their distributions to further mitigate the data distribution gaps.

\begin{figure}[tbp]
\centerline{\includegraphics[width=\linewidth]{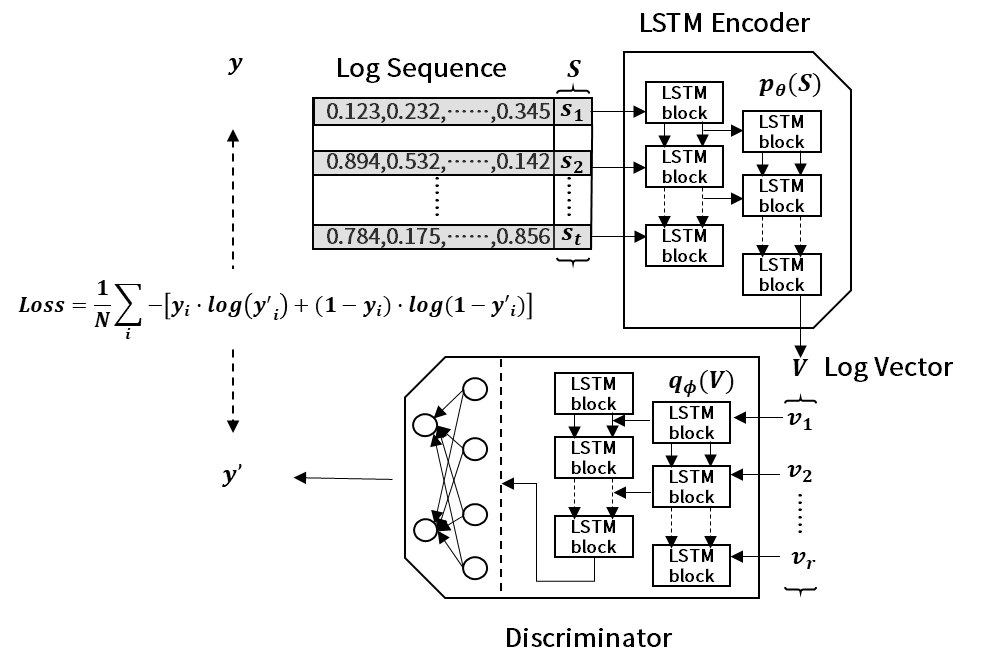}}
\caption{The overview of encoder.}
\label{fig:encoder}
\end{figure}

Figure \ref{fig:contrastive learning} illustrates the encoding process, wherein we consider the normal log sequences from the source system and the target system as one class (Class 0), while treating the anomalous log sequences from the source and target systems as another class (Class 1). By employing contrastive learning, we aim to minimize the intra-class distance and maximize the inter-class distance. Specifically, the input to the encoder is a log sequence $S=\{s_1,s_2,...,s_t\}$, which has undergone word embedding processing. Here, $s$ represents a embedded log event, and $t$ denotes the time window size. The class of each log sequence is annotated by $y\in\{0,1\}$, when $y=0$, it indicates that the log sequence $S$ belongs to Class 0, and when $y=1$, it signifies that the log sequence $S$ belongs to Class 1. We employ the encoder to encode the log sequence $S$ into a log vector $V=\{v_1,v_2,...,v_r\}$, effectively modeling the distribution $V = p_\theta(S)$, where $\theta$ denotes the parameters of the encoder and $r$ denotes the encoding dimensionality. During training, we construct a discriminator to identify the class to which the encoded result $V$ belongs, modeling the distribution $y' = q_\phi(V)$. The objective of the discriminator is to correctly identify the class to which the encoded log vector belongs. Our intuition is that if the encoded results $V$ can concentrate in the feature space based on their respective classes, they should be more easily distinguishable by the discriminator. 
As illustrated in Figure \ref{fig:encoder}, the encoder's architecture comprises two LSTM layers and a fully connected layer. The log vector $V$ serves as the input, and the output being $y' \in [0,1]$, where $y'$ and $1-y'$ represent the probabilities of $V$ belonging to Class 0 and Class 1, respectively. We utilize cross-entropy loss to assess the encoding performance, given by 
\begin{equation}
\mathcal{L} = \frac{1}{N}\sum_{i}^{N}-\left [y_i \cdot \log(y_i') + (1-y_i) \cdot \log(1-y_i') \right ],
\label{equ1}
\end{equation}
In this context, $N$ represents the number of training log sequences. We employ $L$ to update our parameters $\theta$ and $\phi$. Specifically, our training objective is to minimize the following objective function:

% \begin{equation}
% \small
% \begin{split}
%    &\theta,\phi = \underset{\theta,\phi}{argmin}\{\mathcal{L}\}, \\
%    &= \underset{\theta,\phi}{argmin}\{\frac{1}{N}\sum_{i}^{N}-\left [y_i \cdot \log(y_i') + (1-y_i) \cdot \log(1-y_i') \right ]\},\\
%    &= \underset{\theta,\phi}{argmin}\{\frac{1}{N}\sum_{i}^{N}-\left [y_i \cdot \log(q_\phi(p_\theta(S_i))) \\
%    &\qquad\qquad\qquad\qquad\qquad+ (1-y_i) \cdot \log(1-q_\phi(p_\theta(S_i))) \right]\},
% \end{split}
% \label{equ2}
% \end{equation}

\begin{equation}
\small
\begin{split}
    &\theta,\phi = \underset{\theta,\phi}{argmin}\{\mathcal{L}\}, \\
    &= \underset{\theta,\phi}{argmin}\{\frac{1}{N}\sum_{i}^{N}-\left[y_i \cdot \log(y_i') + (1-y_i) \cdot \log(1-y_i') \right]\},\\
    &= \underset{\theta,\phi}{argmin}\{\frac{1}{N}\sum_{i}^{N}-\left[y_i \cdot \log(q_\phi(p_\theta(S_i))) \right. \\
    &\qquad\qquad\qquad\qquad\qquad \left.\left.+ (1-y_i) \cdot \log(1-q_\phi(p_\theta(S_i))) \right]\right\},
\end{split}
\label{equ2}
\end{equation}

After training, we utilize the encoder to encode the log sequence $S$, and employ the encoded log vector $V$ to accomplish the active domain adaptation process. 
%A detailed explanation of this process will be provided in the section \ref{sec:ActiveDomainAdaptation}.

\subsection{Active Domain Adaptation}
\label{sec:ActiveDomainAdaptation}
After encoding, we utilize the log vectors from the source system to train a base anomaly detection model, referred to as Classifier(Source). Subsequently, we consistently optimize the Classifier(Source) using online human labels from the target systems. In order to reduce the human labeling efforts, we employ an active learning approach, selecting subsets that provide the most useful information. In our view, the most valuable log vectors would be those that have not been learned by the model. The information within them should differ from the historically labeled log vectors from the source system and those already labeled during the active learning process. Accordingly, we employ two selection strategies to meet this requirement: free energy-based sampling and uncertainty-based sampling, as shown in Algorithm \ref{algorithm:sample selection}. 

\begin{algorithm}
\caption{Sample Selection}
\label{algorithm:sample selection}
\begin{algorithmic}[1]
    \State \textbf{Input:} The pool of unlabeled log vectors from the target system $\mathcal{V}_t$, Classifier(source) $E$
    \Function{CalculateFreeEnergy}{$V$,$E$} 
        \State $\mathcal{F}(V) \gets$ $-log\sum_{l\in\{0,1\}}exp(-E(V,l))$ \Comment{Calculate Free Energy}
        \State \textbf{return} $\mathcal{F}(V)$ \Comment{Output: Free Energy for vector $V$}
    \EndFunction
    
    \Function{CalculateUncertainty}{$V$,$E$} 
        \State $P(V,0) \gets$ $\dfrac{E(V,0)}{\sum_{l'\in\{0,1\}}E(V,l)}$
        \State $P(V,1) \gets$ $\dfrac{E(V,1)}{\sum_{l'\in\{0,1\}}E(V,l)}$
        \State $U(V) \gets \left\vert P(V,0)-P(V,1) \right\vert$ \Comment{Calculate uncertainty}
        \State \textbf{return} $U(V)$ \Comment{Output: Uncertainty for vector $V$}
    \EndFunction

    \Function{SampleSelection}{$\mathcal{V}_t$,$E$,$\Delta_1$,$\Delta_2$}
        \State $\mathcal{F}_{set} \gets \emptyset$ 
        \State $U_{set} \gets \emptyset$ 
        \For{$V_i$ \textbf{in} $\mathcal{V}$} \Comment{Iterate through log vectors}
            \State $\mathcal{F}_{set}[i] \gets $ CalculateFreeEnergy($V_i$,$E$)
            \State $ U_{set}[i] \gets $ CalculateUncertainty($V_i$,$E$)
        \EndFor
        \State $\mathcal{V}_t^1 \gets$ Sampling the maximum free energy though $\mathcal{F}_{set}$ within $\mathcal{V}$ at a sampling rate of $\Delta_1$
        \State $\mathcal{V}_t^2 \gets$ Sampling the maximum uncertainty though $U_{set}$ within $\mathcal{V}_t^1$ at a sampling rate of $\Delta_2$
        \State \textbf{return} $\mathcal{V}_t^2$ 
    \EndFunction
    
    \State \textbf{Output:} Sampling subset $\mathcal{V}_t^2$ utilized for manual labeling
\end{algorithmic}
\end{algorithm}

In the phase based on free energy-based sampling, we select the log vectors that differ from the historically labeled log vectors in source system. Specifically, we model the log vectors from the source system as a Gaussian distribution. In this Gaussian distribution, log vectors with high free energy imply that they deviate from the center and reside at the boundary of the distribution. \added{In other words, the rationale behind free energy-based sampling is to use Gaussian distributions to model log vector distributions. By labeling and subsequently training on log vectors that fall within the distribution gap, \method can effectively reduce the distribution gap between the source and target systems, thereby facilitating model transfer.}
Therefore, we selecting the target system's log vectors that has the highest free energy in Gaussian distribution for human labeling. These log vectors differ from historical log vectors in source system. Moreover, this subset of log vectors exhibits a distribution distinct from that of the source system, occupying the data distribution gap between the source and target systems. Learning from this specific set of data facilitates further reduction in the distribution gap between the systems. Inspired by EADA~\cite{EADA}, we use the energy-based model (EBM)~\cite{lecun2006tutorial} to build the Gaussian distribution for source system log vectors. Generally, we employ an EBM as Classifier(source) and train it using log vectors $V$ from the source system alongside corresponding labels $l \in \{0,1\}$. Here, $l=0$ signifies a normal log vector, and $l=1$ represents an anomalous log vector. Classifier(source) takes log vectors $V$ as input and produces the energy $E(V,0)$ and $E(V,1)$ for the normal and anomalous distributions, respectively. The training objective of Classifier(source) aims to minimize the energy assigned for correct classification. In other words, for an input log vector $V$ and label $l$, we aim to satisfy the condition:
\begin{equation}
l = argmin_{l'\in\{0,1\}}E(V,l')
\label{equ3},
\end{equation}
After training, we can employ Energy-Based Models (EBMs) to model the joint distribution, $p(V,l)$, of the input vector $V$ and the corresponding label $l$:
\begin{equation}
\begin{split}
   p(V,l)&=exp(-E(V,l))/Z, \\
   Z&=\sum_{V\in\mathcal{V}}\sum_{l\in\{0,1\}}exp(-E(V,l)),
\end{split}
\label{equ4}
\end{equation}
Correspondingly, the marginal distribution of $V$ can be obtained as follows:
\begin{equation}
p(V)=\sum_{l\in\{0,1\}}p(V,l)=\sum_{l\in\{0,1\}}exp(-E(V,l))/Z,
\label{equ5}
\end{equation}
Simultaneously, $p(V)$ can be represented by the free energy $\mathcal{F}(V)$ of $V$:
\begin{equation}
p(V)=\dfrac{exp(-\mathcal{F}(V)}{\sum_{V\in\mathcal{V}}exp(-\mathcal{F}(x))} ,
\label{equ6}
\end{equation}
So, according to Equations \ref{equ5} and \ref{equ6}, we can calculate the free energy of the Gaussian distribution for the input log vector $V$:
\begin{equation}
\mathcal{F}(V)=-log\sum_{l\in\{0,1\}}exp(-E(V,l)),
\label{equ7}
\end{equation}
According to Equation \ref{equ7}, we calculate the free energy $\mathcal{F}(V)$ within the target system's unlabeled log vectors, and by selecting a subset of log vectors with high free energies, we accomplish the free energy-based sampling part.

\added{The rationale behind uncertainty-based sampling is to label log vectors for which the model exhibits the greatest difficulty in making accurate predictions. By learning through these challenging log vectors, \method can improve the performance of the anomaly detection model on the target system.}
In the uncertainty-based sampling phase, we select log vectors that are distinct from those already labeled in the active learning process. More precisely, we choose the target system's log vectors positioned at the boundary between normal and anomalous classifications for manual labeling. These instances signify scenarios the model hasn't encountered previously, as the model would otherwise confidently classify them into their respective categories. Specifically, for a log vector $V$ in the pool of unlabeled log vectors, the Classifier(source) can provide the probabilities $P(V,0)$ and $P(V,1)$ for $V$ belonging to normal log sequences and anomalous log sequences, respectively:
\begin{equation}
\begin{split}
   P(V,0) &= \dfrac{E(V,0)}{\sum_{l'\in\{0,1\}}E(V,l)},\\
   P(V,1) &= \dfrac{E(V,1)}{\sum_{l'\in\{0,1\}}E(V,l)},\\
\end{split}
\label{equ8}
\end{equation}
If $P(V,0)$ and $P(V,1)$ are closer, it indicates that the model lacks confidence in classifying the log vector $V$. Therefore, we calculate the absolute difference between $P(V,0)$ and $P(V,1)$ to determine the uncertainty $U(V)$ of the model for the log vector $V$:
\begin{equation}
   U(V) = \left\vert P(V,0)-P(V,1) \right\vert,
\label{equ9}
\end{equation}

We make selections based on log vectors' uncertainty and free energy, and employ fine-tuning of the selected vectors for the Classifier(source). This adaptation aims to enhance its performance in the target system's log anomaly detection task. Ultimately, we obtain a Classifier(target) with improved generalization capabilities. 
% The detailed sample selection algorithm is provided in Appendix \ref{algorithm:sample selection}.
\section{Experiment}
\label{sec:experiment}
Ours experiments focus on the following research questions (RQS):
\begin{itemize}
\item \textbf{RQ1:} How does \method performs in terms of effectiveness?
\item \textbf{RQ2:} How does the varying number of labels impact the effectiveness of the \method?
\item \textbf{RQ3:} Does each main component contribute to \method?
\end{itemize}

\subsection{Dataset and Experiment Settings}
\label{sec:dataset}
We conducted our experiments on three public log datasets from Loghub~\cite{zhu2023loghub}: BGL, Thunderbird, and Zookeeper. 
\textbf{BGL} refers to an open dataset comprising logs gathered from a BlueGene/L supercomputer system~\cite{oliner2007supercomputers}. \textbf{Thunderbird} contains logs obtained from a Thunderbird supercomputer system~\cite{oliner2007supercomputers}, with 9,024 processors and 27,072GB memory. \textbf{ZooKeeper} contains logs amassed by aggregating data from the ZooKeeper service~\cite{hunt2010zookeeper} within a lab at CUHK. 
These log datasets originate from three entirely distinct systems. 
We designate one dataset as the source system and another as the target for experimentation, resulting in a total of six combinations. We split each set of data into training and testing sets with a 7:3 ratio. \added{To prevent data leakage, we strictly split the training and testing sets based on time, ensuring a certain time gap between them to avoid any data leakage.} When the dataset serves as the target system, the training set becomes the sample pool for selection. The detailed contents of each dataset are outlined in Table \ref{tab:dataset}. 

Our code can be accessed via \url{https://anonymous.4open.science/r/LogAction-B821}.
The experiments were carried out on a Linux server equipped with an Intel(R) Xeon(R) Gold 6126T 2.60GHz CPU, 256GB of memory, and four RTX A4000 GPUs with 128GB of GPU memory. \added{During training, we use the Adam optimizer with a learning rate of \(1 \times 10^{-3}\). The number of epochs is set to 60, and the batch size is 512. The dimensional settings of the model layers, such as the hidden size of the LSTM in the encoder, are adopted from previous seminal works~\cite{deeplog, metalog}. These parameters remain fixed throughout all experiments without modification. In the active domain adaptation process, the energy alignment weight is set to 0.01, the energy-based sampling rate to 0.1, and the active ratio to 0.01.}

\begin{table}[htbp]
  \renewcommand\arraystretch{1.5}
  \small
  \begin{center}
    \caption{Summary of the BGL , ThunderBird, and Zookeeper dataset}
    \label{tab:dataset}
    \scalebox{0.9}{
    \begin{tabular}{cccc@{\hspace{0.1mm}}cc}
    \hline
    \multirow{2}{*}{\textbf{Dataset}}
    & \multicolumn{2}{c}{\textbf{Training Set / Sample 
 Pool}} 
    &{}
    & \multicolumn{2}{c}{\textbf{Testing Set}} \\
    \cline{2-3}
    \cline{5-6}
    & \textbf{\#Normal} & \textbf{\#Anomalous} 
    &{}
    & \textbf{\#Normal} & \textbf{\#Anomalous} \\
    \hline
    BGL & 26367 & 8633 &{} & 11301 & 3699 \\
    ThunderBird & 33205 & 1795 &{} & 14231 & 769 \\
    Zookeeper & 25412 & 614 &{} & 10892 & 262 \\
    \hline
    \end{tabular}
    }
  \end{center}
\end{table}

\begin{table}[htbp]
    \renewcommand\arraystretch{1.6}
    \centering
    % 标题单独标记
    \caption{\added{Hyperparameters}}
    \begin{tabular}{ccc}
        \hline
        % 表头单独标记
        \textbf{\added{Parameter Category}} & \textbf{\added{Parameter Name}} & \textbf{\added{Value}} \\ \hline
        % 第一类参数
        \multirow{3}{*}{\added{Training Configuration}} & \added{epoch} & \added{60} \\
        & \added{batch size} & \added{512} \\ 
        & \added{learning rate} & \added{$10^{-3}$} \\ \hline
        % 第二类参数
        \multirow{2}{*}{\added{Encoder}} & \added{LSTM layer} & \added{2} \\
        & \added{LSTM hidden size} & \added{512} \\ \hline
        % 第三类参数
        \multirow{3}{*}{\added{Anomaly Detection Model}} & \added{Input Size} & \added{512} \\
        & \added{Hidden size} & \added{64} \\ 
        & \added{Layer size} & \added{64} \\ \hline
        % 第四类参数
        \multirow{3}{*}{\added{Active Domain Adaptation}} 
        & \added{Energy align weight} & \added{0.01} \\
        & \added{First sample ratio} & \added{0.1} \\
        & \added{Active ratio} & \added{0.01} \\ \hline
    \end{tabular}
    \label{tab:hyperparameters}
\end{table}

\subsection{Competitors and Implementation Details}
\label{sec:competitors}
For RQ1, we opted for several state-of-the-art transfer-learning-based methods, active-learning-based methods, and some deep learning methods for comparison against \method. For RQ2, we investigated the sensitivity of \method to varying levels of human labeling by setting different quantities of manual labels. 
For RQ3, we delve into the specific roles of \replaced[id=replace]{each main component}{transfer learning and active learning }within \method. We conducted a dissection of \method by separately dismantling its transfer learning, active learning \added{and tow sampling strategy} components, resulting in \replaced[id=replace]{four}{two} variants.
Through comparison, we explored the individual impacts of each component. Below is an introduction to each of the competitors. 

\subsubsection{Transfer-learning-based methods}
\begin{itemize}
\item \textbf{LogTransfer}~\cite{logtransfer} is a supervised transfer learning method. It involves training a model on the source system using a shared network and fine-tuning the first half of the network with the target system logs to complete the transfer learning process.
\item \textbf{LogTAD}~\cite{logtad} is an unsupervised transfer learning method. It learns the hypersphere center of normal logs from the source system's logs and utilizes normal logs from the target system to transfer hypersphere center, thereby achieving the purpose of model migration.
\item \textbf{MetaLog}~\cite{metalog} is a generalizable cross-system anomaly detection approach, it utilizes a meta-learning paradigm to enhance the model's generalization ability, constructing anomaly detection models for new systems with limited labels.
\end{itemize}

\subsubsection{Active-learning-based methods}
\begin{itemize}
\item \textbf{ACLog}~\cite{aclog} utilizes normal logs to train an unsupervised model and employs active learning to select logs within a 'fuzzy' window for human labeling. These labeled logs are used to denoise and enhance the original training set, thereby boosting the performance of the unsupervised model.
\end{itemize}

\subsubsection{Deep learning methods}
\begin{itemize}
\item \textbf{LogCluster}~\cite{vaarandi2015logcluster} organizes and clusters historical logs to extract information aimed at identifying anomalous logs.
\item \textbf{DeepLog}~\cite{deeplog} employs LSTM networks to predict the concluding log event in a sequence of normal logs. It detects anomalies by assessing the variance between the predicted log event and the actual log event that occurs.
\end{itemize}

\subsubsection{Variants of \method}
\begin{itemize}
\item \textbf{${\method}_{wt}$} (\method without transfer learning) dismantles the transfer-learning component, omitting the utilization of logs from the source system to train the base model. Instead, it directly employs active learning to select samples to train the anomaly detection model on the target system. ${\method}_{wt}$ is utilized to explore the role of the transfer-learning component.
\item \textbf{${\method}_{wa}$} (\method without active learning) replaces the active sample selection strategy with a random sample selection of equivalent size. ${\method}_{wa}$ is employed to explore the function of the active learning component.
\item \textbf{\added{${\method}_{wu}$}} \added{(\method without uncertainty-based sampling) is employed to explore the function of the uncertainty-based sampling strategy.}
\item \textbf{\added{${\method}_{we}$}} \added{(\method without free energy-based sampling) is employed to explore the function of the energy-based sampling strategy.}
\end{itemize}
We consistently used the same labeled quantity for the target system whenever required (2\% relative to the sample pool). For models with sample selection strategies (such as ACLog and ${\method}_{wt}$), we conducted two rounds of selection, with 1\% of the samples chosen in each round. For methods lacking sample selection strategies (such as LogTAD and ${\method}_{wa}$), we utilized random sample selection to assign labels, and the ultimate outcomes were derived by averaging across 5 random iterations. For RQ2, we examined the influence of manual labeling quantities ranging from 0\% to \replaced[id=replace]{5\%}{10\%} (at intervals of \replaced[id=replace]{0.5\%}{1\%}) on six combinations of datasets using the \method. The main adjustment involved varying the quantity of human labeling by augmenting the selection rounds, wherein each round involved the selection of \replaced[id=replace]{0.5\%}{1\%} of the samples.

\subsection{Measurements}
\label{sec:measurements}
To evaluate the performance of \method in target test set, we utilize three assessment metrics: \emph{Precision}, \emph{Recall}, and \emph{F1-score}. These metrics rely on four key parameters: True Positive (TP), False Positive (FP), True Negative (TN), and False Negative (FN). The \emph{Precision} metric measures the accuracy of anomaly detection within our system. It's calculated as the ratio of TP to the sum of TP and FP, expressed as $\frac{TP}{TP+FP}$. Conversely, the \emph{Recall} metric assesses the system's capability to detect all anomalous log sequences. It's computed as the ratio of TP to the sum of TP and FN, represented by $\frac{TP}{TP+FN}$. The \emph{F1-score} represents the harmonic mean of precision and recall, providing a balanced evaluation incorporating both precision and recall. Its computation follows this formula: $\frac{2\cdot(Precision\cdot Recall)}{Precision+Recall}$.

\begin{table*}[htbp]
\small
  \renewcommand\arraystretch{1.2}
  \begin{center}
    \caption{Comparison results with baseline methods for Cross-system Anomaly Detection}
    \label{tab:rq1}
    \scalebox{1.1}{
    \begin{tabular}{ccccccccccccc}
    \hline
    \multirow{2}{*}{\textbf{Method}}
    & \multicolumn{3}{c}{\textbf{$\text{ThunderBird} \to \text{BGL}$}}
    & {}
    & \multicolumn{3}{c}{\textbf{$\text{Zookeeper} \to \text{BGL}$}}
    & {}
    & \multicolumn{3}{c}{\textbf{$\text{BGL} \to \text{Zookeeper}$}} 
    \\
    \cline{2-4}
    \cline{6-8}
    \cline{10-12}
    & \textbf{F1} & \textbf{Precision} & \textbf{Recall}
    &{}
    & \textbf{F1} & \textbf{Precision} & \textbf{Recall}
    &{}
    & \textbf{F1} & \textbf{Precision} & \textbf{Recall}
    \\
    \hline
    LogCluster
    & \multicolumn{1}{c}{75.35\%} & \multicolumn{1}{c}{69.63\%} & \multicolumn{1}{c}{82.10\%}
    &{}
    & \multicolumn{1}{c}{75.35\%} & \multicolumn{1}{c}{69.63\%} & \multicolumn{1}{c}{82.10\%}
    &{}
    & \multicolumn{1}{c}{22.65\%} & \multicolumn{1}{c}{17.87\%} & \multicolumn{1}{c}{30.92\%} 
    \\
    DeepLog
    & \multicolumn{1}{c}{82.02\%} & \multicolumn{1}{c}{72.66\%} & \multicolumn{1}{c}{94.16\%}
    &{}
    & \multicolumn{1}{c}{82.02\%} & \multicolumn{1}{c}{72.66\%} & \multicolumn{1}{c}{94.16\%}
    &{}
    & \multicolumn{1}{c}{29.04\%} & \multicolumn{1}{c}{26.48\%} & \multicolumn{1}{c}{32.14\%} 
    \\
    MetaLog
    & \multicolumn{1}{c}{90.80\%} & \multicolumn{1}{c}{90.24\%} & \multicolumn{1}{c}{91.37\%}
    &{}
    & \multicolumn{1}{c}{89.33\%} & \multicolumn{1}{c}{86.22\%} & \multicolumn{1}{c}{92.68\%}
    &{}
    & \multicolumn{1}{c}{60.94\%} & \multicolumn{1}{c}{65.21\%} & \multicolumn{1}{c}{57.20\%} 
    &{}
    \\
    LogTransfer
    & \multicolumn{1}{c}{91.54\%} & \multicolumn{1}{c}{89.03\%} & \multicolumn{1}{c}{94.19\%}
    &{}
    & \multicolumn{1}{c}{94.39\%} & \multicolumn{1}{c}{92.66\%} & \multicolumn{1}{c}{96.19\%}
    &{}
    & \multicolumn{1}{c}{57.66\%} & \multicolumn{1}{c}{70.33\%} & \multicolumn{1}{c}{48.85\%} 
    \\
    LogTAD
    & \multicolumn{1}{c}{89.20\%} & \multicolumn{1}{c}{88.32\%} & \multicolumn{1}{c}{90.10\%}
    &{}
    & \multicolumn{1}{c}{90.20\%} & \multicolumn{1}{c}{90.96\%} & \multicolumn{1}{c}{89.45\%}
    &{}
    & \multicolumn{1}{c}{50.96\%} & \multicolumn{1}{c}{48.14\%} & \multicolumn{1}{c}{54.12\%} 
    \\
    ACLog
    & \multicolumn{1}{c}{90.37\%} & \multicolumn{1}{c}{95.34\%} & \multicolumn{1}{c}{85.90\%}
    &{}
    & \multicolumn{1}{c}{90.37\%} & \multicolumn{1}{c}{95.34\%} & \multicolumn{1}{c}{85.90\%}
    &{}
    & \multicolumn{1}{c}{40.50\%} & \multicolumn{1}{c}{57.38\%} & \multicolumn{1}{c}{31.29\%} 
    \\
    \rowcolor{gray!20} \textbf{\method}
    & \multicolumn{1}{c}{\textbf{96.03\%}} & \multicolumn{1}{c}{\textbf{96.21\%}} & \multicolumn{1}{c}{\textbf{95.84\%}}
    &{}
    & \multicolumn{1}{c}{\textbf{97.46\%}} & \multicolumn{1}{c}{\textbf{97.06\%}}
    & \multicolumn{1}{c}{\textbf{97.87\%}}
    &{}
    & \multicolumn{1}{c}{\textbf{80.66\%}} & \multicolumn{1}{c}{\textbf{99.86\%}} & \multicolumn{1}{c}{\textbf{67.65\%}} 
    \\
    \hline
    \end{tabular}
    }
  \end{center}
  \begin{center}
  \scalebox{1.1}{
    \begin{tabular}{cccccccccccccccc}
    \hline
    \multirow{2}{*}{\textbf{Method}}
    & \multicolumn{3}{c}{\textbf{$\text{BGL} \to \text{ThunderBird}$}}
    & {}
    & \multicolumn{3}{c}{\textbf{$\text{Zookeeper} \to \text{ThunderBird}$}}
    & {}
    & \multicolumn{3}{c}{\textbf{$\text{ThunderBird} \to \text{Zookeeper}$}} 
    \\
    \cline{2-4}
    \cline{6-8}
    \cline{10-12}
    & \textbf{F1} & \textbf{Precision} & \textbf{Recall}
    &{}
    & \textbf{F1} & \textbf{Precision} & \textbf{Recall}
    &{}
    & \textbf{F1} & \textbf{Precision} & \textbf{Recall}
    \\
    \hline
    LogCluster
    & \multicolumn{1}{c}{71.23\%} & \multicolumn{1}{c}{63.54\%} & \multicolumn{1}{c}{81.03\%}
    &{}
    & \multicolumn{1}{c}{71.23\%} & \multicolumn{1}{c}{63.54\%} & \multicolumn{1}{c}{81.03\%}
    &{}
    & \multicolumn{1}{c}{22.65\%} & \multicolumn{1}{c}{17.87\%} & \multicolumn{1}{c}{30.92\%} 
    \\
    DeepLog
    & \multicolumn{1}{c}{34.43\%} & \multicolumn{1}{c}{54.19\%} & \multicolumn{1}{c}{25.23\%}
    &{}
    & \multicolumn{1}{c}{34.43\%} & \multicolumn{1}{c}{54.19\%} & \multicolumn{1}{c}{25.23\%}
    &{}
    & \multicolumn{1}{c}{29.04\%} & \multicolumn{1}{c}{26.48\%} & \multicolumn{1}{c}{32.14\%} 
    \\
    MetaLog
    & \multicolumn{1}{c}{82.87\%} & \multicolumn{1}{c}{80.98\%} & \multicolumn{1}{c}{84.85\%}
    &{}
    & \multicolumn{1}{c}{83.96\%} & \multicolumn{1}{c}{75.71\%} & \multicolumn{1}{c}{94.22\%}
    &{}
    & \multicolumn{1}{c}{57.60\%} & \multicolumn{1}{c}{62.05\%} & \multicolumn{1}{c}{53.75\%} 
    \\
    LogTransfer
    & \multicolumn{1}{c}{80.60\%} & \multicolumn{1}{c}{75.08\%} & \multicolumn{1}{c}{87.00\%}
    &{}
    & \multicolumn{1}{c}{84.11\%} & \multicolumn{1}{c}{80.74\%} & \multicolumn{1}{c}{87.78\%}
    &{}
    & \multicolumn{1}{c}{55.35\%} & \multicolumn{1}{c}{61.40\%} & \multicolumn{1}{c}{50.38\%} 
    \\
    LogTAD
    & \multicolumn{1}{c}{83.87\%} & \multicolumn{1}{c}{93.45\%} & \multicolumn{1}{c}{76.07\%}
    &{}
    & \multicolumn{1}{c}{79.91\%} & \multicolumn{1}{c}{69.99\%} & \multicolumn{1}{c}{93.11\%}
    &{}
    & \multicolumn{1}{c}{51.89\%} & \multicolumn{1}{c}{58.88\%} & \multicolumn{1}{c}{46.39\%} 
    \\
    ACLog
    & \multicolumn{1}{c}{46.10\%} & \multicolumn{1}{c}{70.01\%} & \multicolumn{1}{c}{34.36\%}
    &{}
    & \multicolumn{1}{c}{46.10\%} & \multicolumn{1}{c}{70.01\%} & \multicolumn{1}{c}{34.36\%}
    &{}
    & \multicolumn{1}{c}{40.50\%} & \multicolumn{1}{c}{57.38\%} & \multicolumn{1}{c}{31.29\%} 
    \\
    \rowcolor{gray!20} \textbf{\method}
    & \multicolumn{1}{c}{\textbf{92.09\%}} & \multicolumn{1}{c}{\textbf{95.16\%}} & \multicolumn{1}{c}{\textbf{89.20\%}}
    &{}
    & \multicolumn{1}{c}{\textbf{95.52\%}} & \multicolumn{1}{c}{\textbf{93.96\%}} & \multicolumn{1}{c}{\textbf{97.14\%}}
    &{}
    & \multicolumn{1}{c}{\textbf{96.27\%}} & \multicolumn{1}{c}{\textbf{99.19\%}} & \multicolumn{1}{c}{\textbf{93.51\%}} 
    \\
    \hline
    \end{tabular}
    }
  \end{center}
\end{table*}

\subsection{Evaluation Results}
\label{sec:evaluation results}
\subsubsection{\textbf{RQ1:} How does \method performs in terms of effectiveness?} 

\ 

We conducted experiments on six combinations of source $\to$ target systems involving \method and Competitors, with the experimental results depicted in Table \ref{tab:rq1}. It is important to note that DeepLog, LogCluster, and ACLog were not trained using labels from the source systems, hence their performance is solely dependent on the target systems. As illustrated in the Table \ref{tab:rq1}, our approach yielded F1 scores surpassing all competitors across all six datasets, demonstrating the superior performance of \method.

\textbf{Compared to the Deep learning methods}, \method requires only 2\% of labeled data to achieve an average F1 score surpassing DeepLog and LogCluster by 39.63\%. While these unsupervised deep learning methods do not rely on any human labeled data, their performance tends to be inadequate due to the lack of learning from historical anomalous logs. This deficiency arises from the continual emergence of new log types alongside system changes. Because these unsupervised methods detect anomalies based on historical normal logs, encountering new log types often leads to false positives even when such logs are normal. Reflected in the result, they often exhibit lower precision (averaging only 51.74\%) due to these false positives. Conversely, \method learns the occurrence of anomalous logs with minimal human labeling efforts to detect anomalies. When encountering logs of new types, the model does not classify them as anomalies because they do not resemble historical anomalous logs. As a result, \method exhibits a higher precision (averaging 96.91\%).

\textbf{Compared to the transfer-learning-based methods}, 
\added{\method outperforms the state-of-the-art transfer learning method MetaLog by 15.42\% in F1 score. MetaLog requires 1\% of anomaly labels for training. Although this reduces the number of labels needed, it imposes constraints on the type of labels. Since anomalies occur infrequently, obtaining 1\% anomaly logs necessitates filtering them from a vast amount of normal logs, which increases labeling costs. In contrast, \method imposes no restrictions on label types and achieves an F1 score of 93.01\% with only 2\% of the total labeled data, further reducing labeling costs.} \method outperforms LogTransfer and LogTAD by 17.20\% on average F1 score across six combined datasets. As a supervised transfer learning approach, LogTransfer suffers from the gap in data distribution between the source and target systems' logs, resulting in poor performance. When the data distribution in the target system's logs is relatively simple (such as BGL or ThunderBird), LogTransfer demonstrates a certain level of generalization capability (averaging 87.66\% F1 score). However, in instances where the distribution of data in the target system logs is intricate, such as in Zookeeper, substantial disparities arise between the data distributions of the source and target systems. Consequently, the models trained on labeled logs from the source system exhibit limited generalization capability, leading to lower F1-scores, averaging at 56.51\%. LogTAD, an unsupervised transfer learning method, is also affected by the lack of historical anomaly log learning (similar to DeepLog and LogCluster), impacting its performance. It still requires some manual labeling to continually update the model. In contrast, \method mitigates the disparity in data distribution by employing active learning to select samples with the highest information content, thus minimizing the need for human labeling efforts.

\textbf{Compared to the active-learning-based method}, \method achieves an average F1 score improvement of 30.60\% over ACLog. ACLog, based on unsupervised methods, employs active learning to continuously select samples for online model updating, thereby enhancing the performance of the unsupervised model while reducing the required amount of human labeling. However, ACLog utilizes manually labeled anomalous samples for dataset denoising without fully exploiting these anomalous samples. On datasets with higher levels of noise, such as BGL, ACLog exhibits a substantial improvement compared to unsupervised anomaly detection methods (with an average F1 increase of 11.96\%). However, on datasets with lower noise levels, like Thunderbird and Zookeeper, the utilization efficiency of samples chosen through active learning is limited. Consequently, ACLog performs poorly on these datasets (average 43.30\% in F1 score). In contrast, \method, based on supervised methods, can comprehensively learn patterns from anomalous logs, resulting in more efficient anomaly detection. Furthermore, \method mitigates the cold-start issue of ACLog by leveraging labels from other mature source systems for training the base model.

\begin{figure*}[htbp]
\small
\centerline{\includegraphics[width=\linewidth]{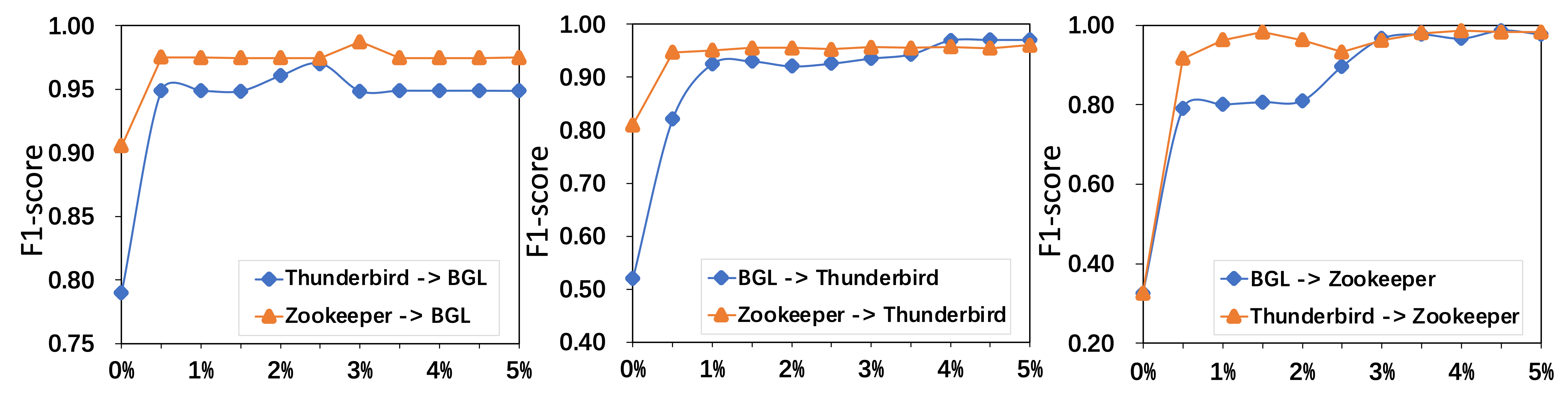}}
\caption{\replaced[id=replace img]{Human labeling efforts}{Human labeling amount and \method F1-score curve}}
\label{fig:human label}
\end{figure*}

\subsubsection{\textbf{RQ2:} How does the varying number of labels impact the effectiveness of the \method?} 
\label{sec:evaluation results rq2}

\ 

We investigated the impact of different levels of manual annotations on \method, and the results are presented in Figure \ref{fig:human label}. Overall, as the volume of labels increased, the performance of \method showed a gradual improvement. Notably, the most significant enhancement was observed between the 0\% and 1\% annotation levels, indicating poor performance of \method when no labels were available. At this stage, Classifier(source) operated without any fine-tuning, solely relying on the logs from the source system to detect anomalies in the target system. Despite encoding, while the log vectors from the source and target systems exhibited similar distributions, Classifier(source) demonstrated some degree of generalization, achieving an average F1 score of 64.10\% on the target system. However, notable distribution gaps persisted between the log vectors of the source and target systems, impeding Classifier(source) from adequately fulfilling the anomaly detection requirements of the target system. Through energy-based sampling and uncertainty-based sampling, \method selected a subset of the most informative samples for human labeling on the target system. These samples contained non-redundant information and encapsulated the distribution gaps between the log vectors of the source and target systems. Following fine-tuning on this subset of log vectors, Classifier(source) transformed into Classifier(target), more suitable for anomaly detection in the target system. Remarkably, a mere 1\% of manual annotations led to an average 22.06\% improvement in the F1 score of the model on the target system. 
\added{
Even with a very limited labeling proportion (0.5\%), \method can achieve an average F1-score of 89.95\%. This demonstrates \method’s effective ability to balance labeling cost and performance. In other words, the amount of labeled logs input to \method can be adjusted according to practical needs. For example, when logs are difficult to label, \method can be initiated with 0.5\% or even less labeled logs. Conversely, if labeling costs are low, 2\% labeled data can be used to achieve the full performance of \method.
}

Furthermore, as depicted in the Figure \ref{fig:human label}, once the level of manual annotations reached a certain threshold, the model's performance plateaued. Additional manual annotations at this stage resulted in minimal improvement as the most valuable samples had already been broadly selected, leaving mostly redundant samples. Further model training on these redundant samples yielded little additional benefit since the model had already captured and learned from the information contained within them.

\begin{figure}[tbp]
\centerline{\includegraphics[width=\linewidth]{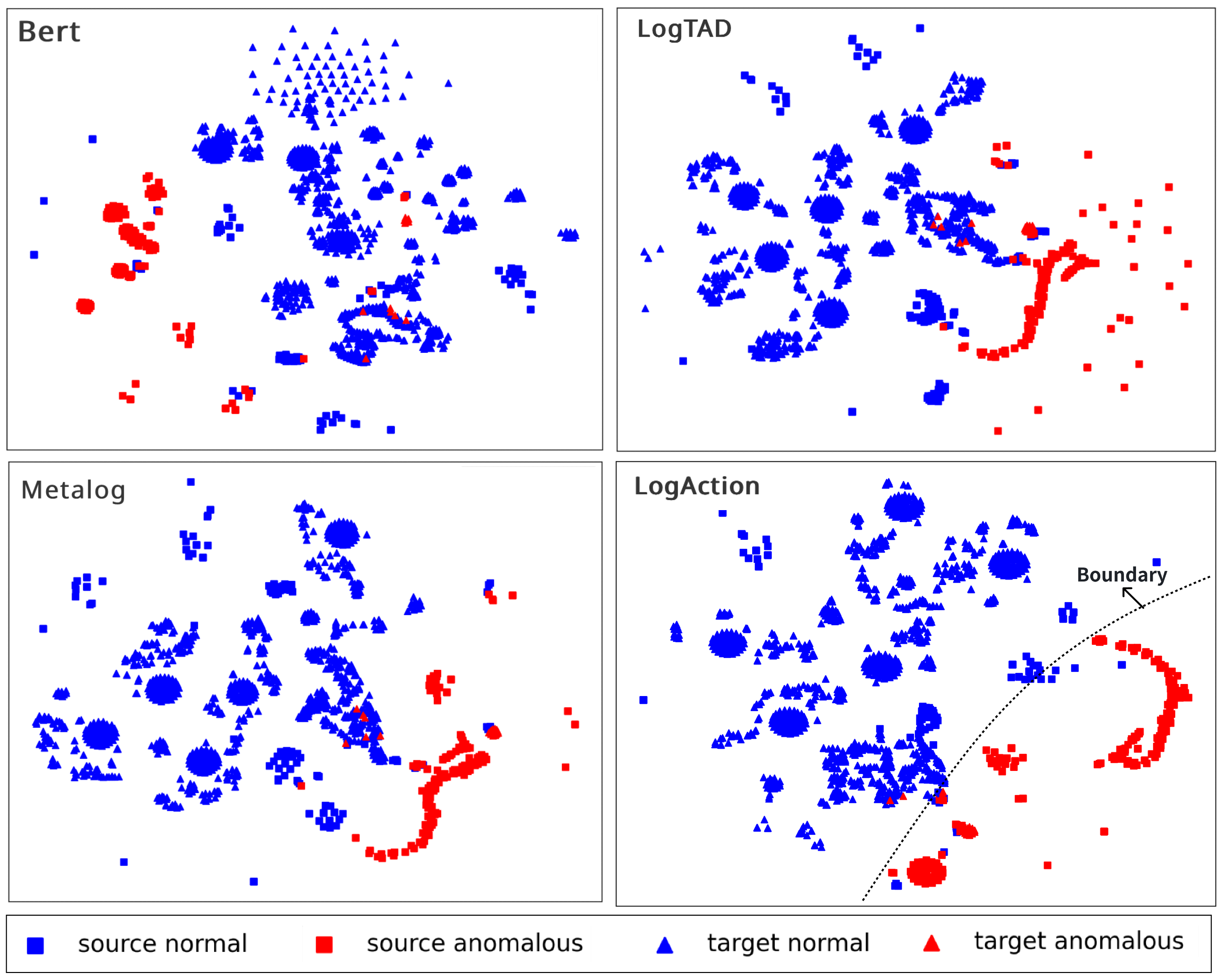}}
\caption{\replaced[id=replace img]{The difference of data distribution before and after encoding in BLG $\to$ ThunderBird.}{The overview of encoder.}}
\label{fig:grap-distribution}
\end{figure}

\begin{table*}[htbp]
\small
  \renewcommand\arraystretch{1.2}
  \begin{center}
    \caption{Ablation of Proposed two Key Components of \method}
    \label{tab:rq3}
    \scalebox{1.1}{
    \begin{tabular}{ccccccccccccc}
    \hline
    \multirow{2}{*}{\textbf{Method}}
    & \multicolumn{3}{c}{\textbf{$\text{ThunderBird} \to \text{BGL}$}}
    & {}
    & \multicolumn{3}{c}{\textbf{$\text{Zookeeper} \to \text{BGL}$}}
    & {}
    & \multicolumn{3}{c}{\textbf{$\text{BGL} \to \text{Zookeeper}$}} 
    \\
    \cline{2-4}
    \cline{6-8}
    \cline{10-12}
    & \textbf{F1} & \textbf{Precision} & \textbf{Recall}
    &{}
    & \textbf{F1} & \textbf{Precision} & \textbf{Recall}
    &{}
    & \textbf{F1} & \textbf{Precision} & \textbf{Recall}
    \\
    \hline
    ${\method}_{wt}$
    & \multicolumn{1}{c}{90.53\%} & \multicolumn{1}{c}{87.55\%} & \multicolumn{1}{c}{93.73\%}
    &{}
    & \multicolumn{1}{c}{90.53\%} & \multicolumn{1}{c}{87.55\%} & \multicolumn{1}{c}{93.73\%}
    &{}
    & \multicolumn{1}{c}{32.60\%} & \multicolumn{1}{c}{50.89\%} & \multicolumn{1}{c}{23.98\%} 
    \\
    ${\method}_{wa}$
    & \multicolumn{1}{c}{89.33\%} & \multicolumn{1}{c}{88.82\%} & \multicolumn{1}{c}{89.86\%}
    &{}
    & \multicolumn{1}{c}{96.39\%} & \multicolumn{1}{c}{96.40\%} & \multicolumn{1}{c}{96.38\%}
    &{}
    & \multicolumn{1}{c}{62.60\%} & \multicolumn{1}{c}{98.87\%} & \multicolumn{1}{c}{45.80\%} 
    &{}
    \\
    \added{${\method}_{wu}$}
    & \multicolumn{1}{c}{\added{92.07\%}} & \multicolumn{1}{c}{\added{90.59\%}} & \multicolumn{1}{c}{\added{93.60\%}}
    &{}
    & \multicolumn{1}{c}{\added{95.32\%}} & \multicolumn{1}{c}{\added{95.29\%}} & \multicolumn{1}{c}{\added{95.35\%}}
    &{}
    & \multicolumn{1}{c}{\added{77.18\%}} & \multicolumn{1}{c}{\added{99.02\%}} & \multicolumn{1}{c}{\added{63.23\%}} 
    \\
    \added{${\method}_{we}$}
    & \multicolumn{1}{c}{\added{91.75\%}} & \multicolumn{1}{c}{\added{93.54\%}} & \multicolumn{1}{c}{\added{90.03\%}}
    &{}
    & \multicolumn{1}{c}{\added{96.91\%}} & \multicolumn{1}{c}{\added{98.03\%}} & \multicolumn{1}{c}{\added{95.81\%}}
    &{}
    & \multicolumn{1}{c}{\added{73.66\%}} & \multicolumn{1}{c}{\added{99.59\%}} & \multicolumn{1}{c}{\added{58.44\%}} 
    \\
    \rowcolor{gray!20} \textbf{\method}
    & \multicolumn{1}{c}{\textbf{96.03\%}} & \multicolumn{1}{c}{\textbf{96.21\%}} & \multicolumn{1}{c}{\textbf{95.84\%}}
    &{}
    & \multicolumn{1}{c}{\textbf{97.46\%}} & \multicolumn{1}{c}{\textbf{97.06\%}}
    & \multicolumn{1}{c}{\textbf{97.87\%}}
    &{}
    & \multicolumn{1}{c}{\textbf{80.66\%}} & \multicolumn{1}{c}{\textbf{99.86\%}} & \multicolumn{1}{c}{\textbf{67.65\%}} 
    \\
    \hline
    \end{tabular}
    }
  \end{center}
  \begin{center}
  \scalebox{1.1}{
    \begin{tabular}{ccccccccccccc}
    \hline
    \multirow{2}{*}{\textbf{Method}}
    & \multicolumn{3}{c}{\textbf{$\text{BGL} \to \text{ThunderBird}$}}
    & {}
    & \multicolumn{3}{c}{\textbf{$\text{Zookeeper} \to \text{ThunderBird}$}}
    & {}
    & \multicolumn{3}{c}{\textbf{$\text{ThunderBird} \to \text{Zookeeper}$}}  
    \\
    \cline{2-4}
    \cline{6-8}
    \cline{10-12}
    & \textbf{F1} & \textbf{Precision} & \textbf{Recall}
    &{}
    & \textbf{F1} & \textbf{Precision} & \textbf{Recall}
    &{}
    & \textbf{F1} & \textbf{Precision} & \textbf{Recall}
    \\
    \hline
    ${\method}_{wt}$
    & \multicolumn{1}{c}{89.32\%} & \multicolumn{1}{c}{\textbf{99.52\%}} & \multicolumn{1}{c}{81.01\%}
    &{}
    & \multicolumn{1}{c}{89.32\%} & \multicolumn{1}{c}{\textbf{99.52\%}} & \multicolumn{1}{c}{81.01\%}
    &{}
    & \multicolumn{1}{c}{32.60\%} & \multicolumn{1}{c}{50.89\%} & \multicolumn{1}{c}{23.98\%} 
    \\
    ${\method}_{wa}$
    & \multicolumn{1}{c}{79.76\%} & \multicolumn{1}{c}{89.82\%} & \multicolumn{1}{c}{71.73\%}
    &{}
    & \multicolumn{1}{c}{90.30\%} & \multicolumn{1}{c}{85.06\%} & \multicolumn{1}{c}{96.23\%}
    &{}
    & \multicolumn{1}{c}{85.66\%} & \multicolumn{1}{c}{80.74\%} & \multicolumn{1}{c}{91.22\%} 
    \\
    \added{${\method}_{wu}$}
    & \multicolumn{1}{c}{\added{85.91\%}} & \multicolumn{1}{c}{\added{93.11\%}} & \multicolumn{1}{c}{\added{79.74\%}}
    &{}
    & \multicolumn{1}{c}{\added{92.31\%}} & \multicolumn{1}{c}{\added{91.28\%}} & \multicolumn{1}{c}{\added{93.36\%}}
    &{}
    & \multicolumn{1}{c}{\added{96.95\%}} & \multicolumn{1}{c}{\added{96.90\%}} & \multicolumn{1}{c}{\added{97.01\%}} 
    \\
    \added{${\method}_{we}$}
    & \multicolumn{1}{c}{\added{86.07\%}} & \multicolumn{1}{c}{\added{92.47\%}} & \multicolumn{1}{c}{\added{80.49\%}}
    &{}
    & \multicolumn{1}{c}{\added{92.98\%}} & \multicolumn{1}{c}{\added{89.69\%}} & \multicolumn{1}{c}{\added{96.51\%}}
    &{}
    & \multicolumn{1}{c}{\added{96.55\%}} & \multicolumn{1}{c}{\added{96.63\%}} & \multicolumn{1}{c}{\added{96.48\%}}
    \\
    \rowcolor{gray!20} \textbf{\method}
    & \multicolumn{1}{c}{\textbf{92.09\%}} & \multicolumn{1}{c}{95.16\%} & \multicolumn{1}{c}{\textbf{89.20\%}}
    &{}
    & \multicolumn{1}{c}{\textbf{95.52\%}} & \multicolumn{1}{c}{93.96\%} & \multicolumn{1}{c}{\textbf{97.14\%}}
    &{}
    & \multicolumn{1}{c}{\textbf{96.27\%}} & \multicolumn{1}{c}{\textbf{99.19\%}} & \multicolumn{1}{c}{\textbf{93.51\%}} 
    \\
    \hline
    \end{tabular}
    }
  \end{center}
\end{table*}

\subsubsection{\textbf{RQ3:} Does each main component contribute to \method?}

\ 

In this section, we perform an ablation study to assess the effectiveness of main components in \method. 
The findings of the ablation study are shown in Table \ref{tab:rq3}. 
In comparison to ${\method}_{wt}$ and ${\method}_{wa}$, \method surpasses them in F1 scores by 21.98\% (${ \method}_{wt}$) and 9.00\% (${ \method}_{wa}$) respectively. Upon removal of the transfer learning, ${\method}_{wt}$ directly select samples in target system without utilizing the logs of the source system to train the basic model, thus encountering a cold start issue. During the initial rounds of free energy-based sampling and uncertainty-based sampling, lacking fundamental distributional knowledge of the target system prevents ${\method}_{wt}$ from selecting the most valuable samples for fine-tuning. In contrast, after encoding, the log vectors from both the source and target systems exhibit similar distributions. \method utilizes the log vectors from the source system to train a basic anomaly detection model, thereby possessing a certain degree of generalization ability to the target system. Consequently, \method accurately models the distribution of the target system, enabling precise active selection and acquisition of the most valuable samples. Compared to ${\method}_{wt}$, the method without the active learning segment, ${\method}_{wa}$, demonstrates higher efficiency. ${\method}_{wa}$ adopts a random selection approach in lieu of the original active learning component, which to some extent can identify valuable samples for fine-tuning. However, its efficacy remains lower than \method since random selection may include redundant samples. Labeling these redundant samples does not enhance model efficiency as the model has already grasped the knowledge they contain. In comparison, \method leverages two sampling methods to avoid generating redundant samples and further alleviate data distribution gaps, thus exhibiting superior performance. \added{We further investigated the contributions of two sampling strategies in active learning. As shown in Table \ref{tab:rq3}, after ablating each sampling method, the model’s performance decreased by an average of 3.05\% and 3.35\%, respectively. This further highlights the importance of both sampling strategies. On one hand, \method employs free energy-based sampling to capture samples located at the distribution boundaries, thereby further reducing the distribution gap between the source and target systems and facilitating model transfer. On the other hand, \method utilizes uncertainty-based sampling to identify samples near the decision boundary between normal and anomalous classes that the model finds difficult to detect, thereby enhancing anomaly detection performance.
}

In addition, we further investigate the role of encoding in the method. For the Encoding component, we utilized the t-SNE dimensionality reduction method to visualize the data distributions of log vectors after encoding. 
\added{We investigated the t-SNE dimensionality reduction results of encoded representations from several different encoders, including BERT, LogTAD, MetaLog, and \method.}
Specifically, the visualization of the BGL $\to$ Thunderbird dataset is depicted in Figure \ref{fig:grap-distribution}, where blue and red represent normal log vectors and anomalous log vectors, respectively. Triangles and circles denote log vectors from the source and target systems. \deleted[id=delete]{Prior to encoding, the distributions of normal and anomalous log vectors from the source and target systems were scattered. In Figure \ref{fig:grap-distribution}, the icons of four distinct colors or shapes cluster separately, indicating dissimilar distributions between the source and target systems. Post encoding, as illustrated in Figure \ref{fig:grap-distribution}, there is a noticeable convergence of normal and anomalous log sequences/vectors from both the source and target systems, indicated by the clustering of various colors in the visualization.} 
\added{Among these, \method’s encoder performs most prominently, as the encoded log vectors from different systems are the most tightly clustered. }
\replaced[id=replace]{After encoding}{Consequently, after encoding,} the data distributions between the source and target systems become analogous, leading to a reduction in the gaps of their distributions. 
\added{\method reduces the distribution gap between the source system and target system logs through contrastive learning, thereby better facilitating model transfer. In comparison, the encoding performance of MetaLog and LogTAD is suboptimal, with many normal and abnormal log vectors intermixed. Moreover, the comparison with the BERT encoder further highlights that the semantic distributions of logs from different systems are inconsistent, making it insufficient to achieve distribution alignment solely by extracting semantics.}

\subsection{\added{Threats to Validity}}
\textbf{\added{Regarding the hyperparameters}}\added{, except for the sensitivity experiments, all other hyperparameters were held constant throughout our experiments. A comprehensive sensitivity analysis of these hyperparameters is therefore lacking. Nonetheless, these parameters are not the primary focus of our study and can be set according to the optimal configurations reported in prior classical works~\cite{deeplog, metalog}. For instance, concerning the LSTM hidden size in the encoder component, the encoder’s generalization capability is predominantly driven by the contrastive learning framework, with the LSTM serving as a standard feature extractor.}

\textbf{\added{Regarding the fixed-size log windows}}\added{, in the BGL, ThunderBird, and Zookeeper datasets, we employ fixed-size log windows to segment logs into log sequences, which may affect model performance. Since other baselines also use fixed-size log windows for these three datasets, to ensure fairness, we adopt the same approach and maintain window sizes consistent with those used in prior studies.}

\textbf{\added{Regarding the labeled logs used in \method,}}\added{ for fairness considerations, we used 2\% labeled logs in the comparative experiments between the baselines and \method. In practice,
obtaining even 2\% labeled logs may incur high costs. However, 2\% labeled logs is not strictly necessary. Striving to balance labeling cost and effectiveness, \method allows the proportion of input labels to be adjusted according to practical needs. As demonstrated in Section \ref{sec:evaluation results rq2}, it maintains strong performance even with as little as 0.5\% of labeled data.
}

\section{Conclusion}
Accurately identifying log labels from a large volume of logs is a highly challenging task, posing difficulties in training anomaly detection models. Existing methods that tackle the scarcity of labeled data primarily involve transfer learning and active learning. Nevertheless, both approaches have their limitations, yet they can be complementary and mutually resolved. In this paper, we propose \method, a novel anomaly detection approach based on active domain adaptation to tackle the issue of limited labeled data. \method combines transfer learning and active learning techniques. On one hand, it employs active learning to selectively label key logs, bridging the gap between distinct systems in transfer learning. On the other hand, \method employs transfer learning, training the base model using labels from the source system, to mitigate the cold-start issue in active learning. Compared to several existing state-of-the-art transfer learning and active learning methods, \method outperforms them by 26.28\% while requiring only 2\% of the amount of human annotation. In the future, we will focus on researching new methods to achieve more precise log-based anomaly detection models with minimal human labeling.

\section*{Acknowledgment}
This research is supported by PKU-ByteDance Research Project.

\bibliographystyle{IEEEtran}
\bibliography{main}

\begin{thebibliography}{10}
\providecommand{\url}[1]{#1}
\csname url@samestyle\endcsname
\providecommand{\newblock}{\relax}
\providecommand{\bibinfo}[2]{#2}
\providecommand{\BIBentrySTDinterwordspacing}{\spaceskip=0pt\relax}
\providecommand{\BIBentryALTinterwordstretchfactor}{4}
\providecommand{\BIBentryALTinterwordspacing}{\spaceskip=\fontdimen2\font plus
\BIBentryALTinterwordstretchfactor\fontdimen3\font minus \fontdimen4\font\relax}
\providecommand{\BIBforeignlanguage}[2]{{%
\expandafter\ifx\csname l@#1\endcsname\relax
\typeout{** WARNING: IEEEtran.bst: No hyphenation pattern has been}%
\typeout{** loaded for the language `#1'. Using the pattern for}%
\typeout{** the default language instead.}%
\else
\language=\csname l@#1\endcsname
\fi
#2}}
\providecommand{\BIBdecl}{\relax}
\BIBdecl

\bibitem{deeplog}
M.~Du, F.~Li, G.~Zheng, and V.~Srikumar, ``Deeplog: Anomaly detection and diagnosis from system logs through deep learning,'' in \emph{Proceedings of the 2017 ACM SIGSAC Conference on Computer and Communications Security}, ser. CCS '17.\hskip 1em plus 0.5em minus 0.4em\relax New York, NY, USA: Association for Computing Machinery, 2017, p. 1285–1298.

\bibitem{loganomaly}
W.~Meng, Y.~Liu, Y.~Zhu, S.~Zhang, D.~Pei, Y.~Liu, Y.~Chen, R.~Zhang, S.~Tao, P.~Sun \emph{et~al.}, ``Loganomaly: Unsupervised detection of sequential and quantitative anomalies in unstructured logs.'' in \emph{IJCAI}, vol.~19, no.~7, 2019, pp. 4739--4745.

\bibitem{icsme2020}
K.~Yin, M.~Yan, L.~Xu, Z.~Xu, Z.~Li, D.~Yang, and X.~Zhang, ``Improving log-based anomaly detection with component-aware analysis,'' in \emph{2020 IEEE International Conference on Software Maintenance and Evolution (ICSME)}, 2020, pp. 667--671.

\bibitem{icse2020}
J.~Kim, V.~Savchenko, K.~Shin, K.~Sorokin, H.~Jeon, G.~Pankratenko, S.~Markov, and C.-J. Kim, ``Automatic abnormal log detection by analyzing log history for providing debugging insight,'' in \emph{Proceedings of the ACM/IEEE 42nd International Conference on Software Engineering: Software Engineering in Practice}, ser. ICSE-SEIP '20.\hskip 1em plus 0.5em minus 0.4em\relax New York, NY, USA: Association for Computing Machinery, 2020, p. 71–80.

\bibitem{he2023unsupervised}
S.~He, T.~Deng, B.~Chen, R.~S. Sherratt, and J.~Wang, ``Unsupervised log anomaly detection method based on multi-feature.'' \emph{Computers, Materials \& Continua}, vol.~76, no.~1, 2023.

\bibitem{llmelog}
M.~He, T.~Jia, C.~Duan, H.~Cai, Y.~Li, and G.~Huang, ``Llmelog: An approach for anomaly detection based on llm-enriched log events,'' in \emph{2024 IEEE 35th International Symposium on Software Reliability Engineering (ISSRE)}, 2024, pp. 132--143.

\bibitem{weakly-supervised-logad}
\BIBentryALTinterwordspacing
------, ``{ Weakly-Supervised Log-Based Anomaly Detection with Inexact Labels via Multi-Instance Learning },'' in \emph{2025 IEEE/ACM 47th International Conference on Software Engineering (ICSE)}.\hskip 1em plus 0.5em minus 0.4em\relax Los Alamitos, CA, USA: IEEE Computer Society, May 2025, pp. 2918--2930. [Online]. Available: \url{https://doi.ieeecomputersociety.org/10.1109/ICSE55347.2025.00189}
\BIBentrySTDinterwordspacing

\bibitem{clslog}
\BIBentryALTinterwordspacing
P.~Xiao, T.~Jia, C.~Duan, M.~He, W.~Hong, X.~Yang, Y.~Wu, Y.~Li, and G.~Huang, \emph{CLSLog: Collaborating Large and Small Models for Log-based Anomaly Detection}.\hskip 1em plus 0.5em minus 0.4em\relax New York, NY, USA: Association for Computing Machinery, 2025, p. 686–690. [Online]. Available: \url{https://doi.org/10.1145/3696630.3728524}
\BIBentrySTDinterwordspacing

\bibitem{plelog}
L.~Yang, J.~Chen, Z.~Wang, W.~Wang, J.~Jiang, X.~Dong, and W.~Zhang, ``Semi-supervised log-based anomaly detection via probabilistic label estimation,'' in \emph{2021 IEEE/ACM 43rd International Conference on Software Engineering (ICSE)}.\hskip 1em plus 0.5em minus 0.4em\relax IEEE, 2021, pp. 1448--1460.

\bibitem{logrobust}
X.~Zhang, Y.~Xu, Q.~Lin, B.~Qiao, H.~Zhang, Y.~Dang, C.~Xie, X.~Yang, Q.~Cheng, Z.~Li, J.~Chen, X.~He, R.~Yao, J.-G. Lou, M.~Chintalapati, F.~Shen, and D.~Zhang, ``Robust log-based anomaly detection on unstable log data,'' in \emph{Proceedings of the 2019 27th ACM Joint Meeting on European Software Engineering Conference and Symposium on the Foundations of Software Engineering}, ser. ESEC/FSE 2019.\hskip 1em plus 0.5em minus 0.4em\relax New York, NY, USA: Association for Computing Machinery, 2019, p. 807–817.

\bibitem{bugidentifier}
W.~Xia, Y.~Li, T.~Jia, and Z.~Wu, ``Bugidentifier: An approach to identifying bugs via log mining for accelerating bug reporting stage,'' in \emph{2019 IEEE 19th International Conference on Software Quality, Reliability and Security (QRS)}, 2019, pp. 167--175.

\bibitem{recurrentfaults}
T.~Reidemeister, M.~A. Munawar, and P.~A. Ward, ``Identifying symptoms of recurrent faults in log files of distributed information systems,'' in \emph{2010 IEEE Network Operations and Management Symposium - NOMS 2010}, 2010, pp. 187--194.

\bibitem{xie2023logrep}
X.~Xie, S.~Jian, C.~Huang, F.~Yu, and Y.~Deng, ``Logrep: Log-based anomaly detection by representing both semantic and numeric information in raw messages,'' in \emph{2023 IEEE 34th International Symposium on Software Reliability Engineering (ISSRE)}.\hskip 1em plus 0.5em minus 0.4em\relax IEEE, 2023, pp. 194--206.

\bibitem{hilog}
T.~Jia, Y.~Li, Y.~Yang, G.~Huang, and Z.~Wu, ``Augmenting log-based anomaly detection models to reduce false anomalies with human feedback,'' in \emph{Proceedings of the 28th ACM SIGKDD Conference on Knowledge Discovery and Data Mining}, 2022, pp. 3081--3089.

\bibitem{logtransfer}
R.~Chen, S.~Zhang, D.~Li, Y.~Zhang, F.~Guo, W.~Meng, D.~Pei, Y.~Zhang, X.~Chen, and Y.~Liu, ``Logtransfer: Cross-system log anomaly detection for software systems with transfer learning,'' in \emph{2020 IEEE 31st International Symposium on Software Reliability Engineering (ISSRE)}.\hskip 1em plus 0.5em minus 0.4em\relax IEEE, 2020, pp. 37--47.

\bibitem{logtad}
X.~Han and S.~Yuan, ``Unsupervised cross-system log anomaly detection via domain adaptation,'' in \emph{Proceedings of the 30th ACM international conference on information \& knowledge management}, 2021, pp. 3068--3072.

\bibitem{metalog}
C.~Zhang, T.~Jia, G.~Shen, P.~Zhu, and Y.~Li, ``Metalog: Generalizable cross-system anomaly detection from logs with meta-learning,'' in \emph{Proceedings of the IEEE/ACM 46th International Conference on Software Engineering}, 2024, pp. 1--12.

\bibitem{aclog}
C.~Duan, T.~Jia, Y.~Li, and G.~Huang, ``Aclog: An approach to detecting anomalies from system logs with active learning,'' in \emph{2023 IEEE International Conference on Web Services (ICWS)}.\hskip 1em plus 0.5em minus 0.4em\relax IEEE, 2023, pp. 436--443.

\bibitem{afalog}
C.~Duan, T.~Jia, H.~Cai, Y.~Li, and G.~Huang, ``Afalog: A general augmentation framework for log-based anomaly detection with active learning,'' in \emph{2023 IEEE 34th International Symposium on Software Reliability Engineering (ISSRE)}.\hskip 1em plus 0.5em minus 0.4em\relax IEEE, 2023, pp. 46--56.

\bibitem{eagerlog}
C.~Duan, T.~Jia, Y.~Yang, G.~Liu, J.~Liu, H.~Zhang, Q.~Zhou, Y.~Li, and G.~Huang, ``Eagerlog: Active learning enhanced retrieval augmented generation for log-based anomaly detection,'' in \emph{ICASSP 2025 - 2025 IEEE International Conference on Acoustics, Speech and Signal Processing (ICASSP)}, 2025, pp. 1--5.

\bibitem{logcae}
P.~Xiao, T.~Jia, C.~Duan, H.~Cai, Y.~Li, and G.~Huang, ``Logcae: An approach for log-based anomaly detection with active learning and contrastive learning,'' in \emph{2024 IEEE 35th International Symposium on Software Reliability Engineering (ISSRE)}, 2024, pp. 144--155.

\bibitem{oliner2007supercomputers}
A.~Oliner and J.~Stearley, ``What supercomputers say: A study of five system logs,'' in \emph{37th annual IEEE/IFIP international conference on dependable systems and networks (DSN'07)}.\hskip 1em plus 0.5em minus 0.4em\relax IEEE, 2007, pp. 575--584.

\bibitem{EADA}
B.~Xie, L.~Yuan, S.~Li, C.~H. Liu, X.~Cheng, and G.~Wang, ``Active learning for domain adaptation: An energy-based approach,'' in \emph{Proceedings of the AAAI Conference on Artificial Intelligence}, vol.~36, no.~8, 2022, pp. 8708--8716.

\bibitem{rai2010domain}
P.~Rai, A.~Saha, H.~Daum{\'e}~III, and S.~Venkatasubramanian, ``Domain adaptation meets active learning,'' in \emph{Proceedings of the NAACL HLT 2010 Workshop on Active Learning for Natural Language Processing}, 2010, pp. 27--32.

\bibitem{han2022loss}
K.~Han, Y.~Kim, D.~Han, and S.~Hong, ``Loss-based sequential learning for active domain adaptation,'' \emph{arXiv preprint arXiv:2204.11665}, 2022.

\bibitem{fu2021transferable}
B.~Fu, Z.~Cao, J.~Wang, and M.~Long, ``Transferable query selection for active domain adaptation,'' in \emph{Proceedings of the IEEE/CVF Conference on Computer Vision and Pattern Recognition}, 2021, pp. 7272--7281.

\bibitem{BART}
M.~Lewis, Y.~Liu, N.~Goyal, M.~Ghazvininejad, A.~Mohamed, O.~Levy, V.~Stoyanov, and L.~Zettlemoyer, ``Bart: Denoising sequence-to-sequence pre-training for natural language generation, translation, and comprehension,'' \emph{arXiv preprint arXiv:1910.13461}, 2019.

\bibitem{AgentFM}
\BIBentryALTinterwordspacing
L.~Zhang, Y.~Zhai, T.~Jia, X.~Huang, C.~Duan, and Y.~Li, ``Agentfm: Role-aware failure management for distributed databases with llm-driven multi-agents,'' in \emph{Proceedings of the 33rd ACM International Conference on the Foundations of Software Engineering}, ser. FSE Companion '25.\hskip 1em plus 0.5em minus 0.4em\relax New York, NY, USA: Association for Computing Machinery, 2025, p. 525–529. [Online]. Available: \url{https://doi.org/10.1145/3696630.3728492}
\BIBentrySTDinterwordspacing

\bibitem{enhancing-web}
\BIBentryALTinterwordspacing
X.~Yang, X.~Huang, C.~Duan, T.~Jia, S.~Dong, Y.~Li, and G.~Huang, ``Enhancing web service anomaly detection via fine-grained multi-modal association and frequency domain analysis,'' in \emph{Companion Proceedings of the ACM on Web Conference 2025}, ser. WWW '25.\hskip 1em plus 0.5em minus 0.4em\relax New York, NY, USA: Association for Computing Machinery, 2025, p. 548–556. [Online]. Available: \url{https://doi.org/10.1145/3701716.3715221}
\BIBentrySTDinterwordspacing

\bibitem{zhang2025thinkflselfrefiningfailurelocalization}
\BIBentryALTinterwordspacing
L.~Zhang, Y.~Zhai, T.~Jia, C.~Duan, S.~Yu, J.~Gao, B.~Ding, Z.~Wu, and Y.~Li, ``Thinkfl: Self-refining failure localization for microservice systems via reinforcement fine-tuning,'' 2025. [Online]. Available: \url{https://arxiv.org/abs/2504.18776}
\BIBentrySTDinterwordspacing

\bibitem{FAMOS}
\BIBentryALTinterwordspacing
C.~Duan, Y.~Yang, T.~Jia, G.~Liu, J.~Liu, H.~Zhang, Q.~Zhou, Y.~Li, and G.~Huang, ``{ Famos: Fault Diagnosis for Microservice Systems Through Effective Multi-Modal Data Fusion },'' in \emph{2025 IEEE/ACM 47th International Conference on Software Engineering (ICSE)}.\hskip 1em plus 0.5em minus 0.4em\relax Los Alamitos, CA, USA: IEEE Computer Society, May 2025, pp. 2613--2624. [Online]. Available: \url{https://doi.ieeecomputersociety.org/10.1109/ICSE55347.2025.00073}
\BIBentrySTDinterwordspacing

\bibitem{soil}
\BIBentryALTinterwordspacing
C.~Duan, F.~Yang, P.~Zhao, L.~Zheng, Y.~Dagli, Y.~Liu, Q.~Lin, and D.~Zhang, ``Soil: Score conditioned diffusion model for imbalanced cloud failure prediction,'' in \emph{Companion Proceedings of the ACM Web Conference 2024}, ser. WWW '24.\hskip 1em plus 0.5em minus 0.4em\relax New York, NY, USA: Association for Computing Machinery, 2024, p. 65–72. [Online]. Available: \url{https://doi.org/10.1145/3589335.3648303}
\BIBentrySTDinterwordspacing

\bibitem{zhang2025adaptivercl}
\BIBentryALTinterwordspacing
L.~Zhang, T.~Jia, K.~Wang, W.~Hong, C.~Duan, M.~He, and Y.~Li, ``Adaptive root cause localization for microservice systems with multi-agent recursion-of-thought,'' 2025. [Online]. Available: \url{https://arxiv.org/abs/2508.20370}
\BIBentrySTDinterwordspacing

\bibitem{logflash}
T.~Jia, Y.~Wu, C.~Hou, and Y.~Li, ``Logflash: Real-time streaming anomaly detection and diagnosis from system logs for large-scale software systems,'' in \emph{2021 IEEE 32nd International Symposium on Software Reliability Engineering (ISSRE)}, 2021, pp. 80--90.

\bibitem{cloudseer}
X.~Yu, P.~Joshi, J.~Xu, G.~Jin, H.~Zhang, and G.~Jiang, ``Cloudseer: Workflow monitoring of cloud infrastructures via interleaved logs,'' \emph{SIGARCH Comput. Archit. News}, vol.~44, no.~2, p. 489–502, mar 2016.

\bibitem{kddcfg}
A.~Nandi, A.~Mandal, S.~Atreja, G.~B. Dasgupta, and S.~Bhattacharya, ``Anomaly detection using program control flow graph mining from execution logs,'' in \emph{Proceedings of the 22nd ACM SIGKDD International Conference on Knowledge Discovery and Data Mining}, ser. KDD '16.\hskip 1em plus 0.5em minus 0.4em\relax New York, NY, USA: Association for Computing Machinery, 2016, p. 215–224.

\bibitem{logsed}
T.~Jia, L.~Yang, P.~Chen, Y.~Li, F.~Meng, and J.~Xu, ``Logsed: Anomaly diagnosis through mining time-weighted control flow graph in logs,'' in \emph{2017 IEEE 10th International Conference on Cloud Computing (CLOUD)}, 2017, pp. 447--455.

\bibitem{icwslogsed}
T.~Jia, P.~Chen, L.~Yang, Y.~Li, F.~Meng, and J.~Xu, ``An approach for anomaly diagnosis based on hybrid graph model with logs for distributed services,'' in \emph{2017 IEEE International Conference on Web Services (ICWS)}, 2017, pp. 25--32.

\bibitem{logdc}
J.~Xu, P.~Chen, L.~Yang, F.~Meng, and P.~Wang, ``Logdc: Problem diagnosis for declartively-deployed cloud applications with log,'' in \emph{2017 IEEE 14th International Conference on e-Business Engineering (ICEBE)}, 2017, pp. 282--287.

\bibitem{ava}
A.~Babenko, L.~Mariani, and F.~Pastore, ``Ava: Automated interpretation of dynamically detected anomalies,'' in \emph{Proceedings of the Eighteenth International Symposium on Software Testing and Analysis}, ser. ISSTA '09.\hskip 1em plus 0.5em minus 0.4em\relax New York, NY, USA: Association for Computing Machinery, 2009, p. 237–248.

\bibitem{beehive}
T.-F. Yen, A.~Oprea, K.~Onarlioglu, T.~Leetham, W.~Robertson, A.~Juels, and E.~Kirda, ``Beehive: Large-scale log analysis for detecting suspicious activity in enterprise networks,'' in \emph{Proceedings of the 29th Annual Computer Security Applications Conference}, ser. ACSAC '13.\hskip 1em plus 0.5em minus 0.4em\relax New York, NY, USA: Association for Computing Machinery, 2013, p. 199–208.

\bibitem{logan}
B.~C. Tak, S.~Tao, L.~Yang, C.~Zhu, and Y.~Ruan, ``Logan: Problem diagnosis in the cloud using log-based reference models,'' in \emph{2016 IEEE International Conference on Cloud Engineering (IC2E)}, 2016, pp. 62--67.

\bibitem{lstm}
A.~Graves, ``Long short-term memory,'' \emph{Supervised sequence labelling with recurrent neural networks}, pp. 37--45, 2012.

\bibitem{su2020active}
J.-C. Su, Y.-H. Tsai, K.~Sohn, B.~Liu, S.~Maji, and M.~Chandraker, ``Active adversarial domain adaptation,'' in \emph{Proceedings of the IEEE/CVF Winter Conference on Applications of Computer Vision}, 2020, pp. 739--748.

\bibitem{drain}
P.~He, J.~Zhu, Z.~Zheng, and M.~R. Lyu, ``Drain: An online log parsing approach with fixed depth tree,'' in \emph{2017 IEEE International Conference on Web Services (ICWS)}, 2017, pp. 33--40.

\bibitem{Glove}
J.~Pennington, R.~Socher, and C.~D. Manning, ``Glove: Global vectors for word representation,'' in \emph{Proceedings of the 2014 conference on empirical methods in natural language processing (EMNLP)}, 2014, pp. 1532--1543.

\bibitem{word2vec}
T.~Mikolov, K.~Chen, G.~Corrado, and J.~Dean, ``Efficient estimation of word representations in vector space,'' \emph{arXiv preprint arXiv:1301.3781}, 2013.

\bibitem{lecun2006tutorial}
Y.~LeCun, S.~Chopra, R.~Hadsell, M.~Ranzato, and F.~Huang, ``A tutorial on energy-based learning,'' \emph{Predicting structured data}, vol.~1, no.~0, 2006.

\bibitem{zhu2023loghub}
J.~Zhu, S.~He, P.~He, J.~Liu, and M.~R. Lyu, ``Loghub: A large collection of system log datasets for ai-driven log analytics,'' in \emph{2023 IEEE 34th International Symposium on Software Reliability Engineering (ISSRE)}.\hskip 1em plus 0.5em minus 0.4em\relax IEEE, 2023, pp. 355--366.

\bibitem{hunt2010zookeeper}
P.~Hunt, M.~Konar, F.~P. Junqueira, and B.~Reed, ``$\{$ZooKeeper$\}$: Wait-free coordination for internet-scale systems,'' in \emph{2010 USENIX Annual Technical Conference (USENIX ATC 10)}, 2010.

\bibitem{vaarandi2015logcluster}
R.~Vaarandi and M.~Pihelgas, ``Logcluster-a data clustering and pattern mining algorithm for event logs,'' in \emph{2015 11th International conference on network and service management (CNSM)}.\hskip 1em plus 0.5em minus 0.4em\relax IEEE, 2015, pp. 1--7.

\end{thebibliography}

\end{document}